\documentclass[sigconf, 9pt]{acmart}
\usepackage[utf8]{inputenc}
\usepackage{natbib}
\usepackage{graphicx}

\usepackage{caption}
\usepackage{subcaption}
\usepackage{xspace}
\usepackage{makecell}
\usepackage{multicol}
\usepackage{multirow}
\usepackage{tikz}

\usepackage[linesnumbered,ruled,vlined]{algorithm2e}
\usepackage{algpseudocode}
\usepackage{amsmath}
\usepackage{paralist}

\SetCommentSty{mycommfont}

\usepackage{threeparttable}

\usepackage{hyperref}

\newcommand*{\circled}[1]{\lower.7ex\hbox{\tikz\draw (0pt, 0pt) circle (.5em) node {\makebox[1em][c]{\small #1}};}}
\newcommand{\code}[1]{\texttt{#1}}

\newcommand{\sysname}[0]{{\sc nnJIT}\xspace}

\newcommand{\eg}{{\it e.g.,}\xspace}
\newcommand{\ie}{{\it i.e.,}\xspace}

\acmYear{2024}\copyrightyear{2024}
\setcopyright{acmcopyright}
\acmConference[MOBISYS '24]{The 22nd Annual International Conference on Mobile Systems, Applications and Services}{June 3--7, 2024}{Minato-ku, Tokyo, Japan}
\acmBooktitle{The 22nd Annual International Conference on Mobile Systems, Applications and Services (MOBISYS '24), June 3--7, 2024, Minato-ku, Tokyo, Japan}
\acmDOI{10.1145/3643832.3661892}
\acmISBN{979-8-4007-0581-6/24/06}

\begin{document}

\title{Empowering In-Browser Deep Learning Inference on Edge Devices with Just-in-Time Kernel Optimizations}

\author{Fucheng Jia}
\authornote{Research interns at Microsoft Research.}
\affiliation{%
   \institution{Central South University\\Microsoft Research}
   \country{}
  }
\email{fuchengjia@csu.edu.cn}

\author{Shiqi Jiang}
\affiliation{%
   \institution{Microsoft Research}
   \country{}
  }
\email{shijiang@microsoft.com}

\author{Ting Cao}
\authornote{Corresponding author.}
\affiliation{%
   \institution{Microsoft Research}
   \country{}
  }
\email{ting.cao@microsoft.com}

\author{Wei Cui}
\affiliation{%
   \institution{Microsoft Research}
   \country{}
  }
\email{weicu@microsoft.com}

\author{Tianrui Xia}
\authornotemark[1]
\affiliation{%
   \institution{University of Southern California\\Microsoft Research}
   \country{}
  }
\email{tianruix@usc.edu}

\author{Xu Cao}
\affiliation{%
   \institution{Microsoft Research}
   \country{}
  }
\email{caox@microsoft.com}

\author{Yuanchun Li}
\affiliation{%
   \institution{Institute for AI Industry Research (AIR), Tsinghua University}
   \country{}
   }
\email{liyuanchun@air.tsinghua.edu.cn}

\author{Qipeng Wang}
\authornotemark[1]
\affiliation{%
   \institution{Peking University\\Microsoft Research}
   \country{}
   }
\email{wangqipeng@stu.pku.edu.cn}

\author{Deyu Zhang}
\affiliation{%
   \institution{Central South University}
   \country{}
  }
\email{zdy876@csu.edu.cn}

\author{Ju Ren}
\affiliation{%
   \institution{Tsinghua University}
   \country{}
  }
\email{renju@tsinghua.edu.cn}

\author{Yunxin Liu}
\affiliation{%
   \institution{Institute for AI Industry Research (AIR), Tsinghua University}
   \country{}
   }
\email{liuyunxin@air.tsinghua.edu.cn}

\author{Lili Qiu}
\affiliation{%
   \institution{Microsoft Research}
   \country{}
  }
\email{liliqiu@microsoft.com}

\author{Mao Yang}
\affiliation{%
   \institution{Microsoft Research}
   \country{}
  }
\email{maoyang@microsoft.com}

\renewcommand{\shortauthors}{F. Jia, S. Jiang, T. Cao, W. Cui, T. Xia, X. Cao, Y. Li, Q. Wang, D. Zhang, J. Ren, Y. Liu, L. Qiu, M. Yang}

\begin{abstract}

Web is increasingly becoming the primary platform to deliver AI services onto edge devices, making in-browser deep learning (DL) inference more prominent. Nevertheless, the heterogeneity of edge devices, combined with the underdeveloped state of Web hardware acceleration practices, hinders current in-browser inference from achieving its full performance potential on target devices.

To address this issue, this paper presents the pioneering in-browser inference system, \sysname, which enables just-in-time (JIT) auto-generation of optimized computing kernels for edge devices. \sysname is built upon two novel techniques that significantly reduce kernel search and compilation overhead while improving performance firmly: Tensor-Web Compiling Co-Design lowers compiling costs by around 100$\times$ through eliminating redundant and ineffective compiling passes; Web-Specific Lite Kernel Optimization Space reduces kernel tuning costs by focusing on Web programming requirements and efficient device resource utilization, pruning the optimization space from millions to only dozens.

\sysname \footnote{Code: \url{https://aka.ms/nnjit-web}} is evaluated for modern models, \eg BART, T5, and Llama 2, on a range of edge devices including laptops and smartphones using different browsers and hardware from ARM, Intel, AMD and Nvidia. The results show that \sysname can achieve up to 8.2$\times$ faster within 30 seconds compared to the existing baselines.

\end{abstract}



\maketitle

\section{Introduction}

Web applications that run through a browser are increasingly popular on edge devices, thanks to their notable benefits such as cross-platform compatibility, effortless \emph{click-and-run} deployment, ease of maintenance, and seamless integration between edge and cloud services \cite{LiFG022}.

Presently, there is a trend towards integrating Deep Neural Network (DNN) services directly into Web applications, enabling in-browser inference. Systems that facilitate in-browser inference have been developed, such as ONNX Runtime Web\cite{microsoft2023ort}, TensorFlow.js\cite{google2023tfjs}, WebDNN\cite{Webdnn}, and brain.js\cite{brainjs}. In-browser inference can provide a more responsive user experience and enhanced privacy protection by avoiding round-trips to cloud, as well as save the expense of cloud computing resources. In-browser inference is made viable by continuous advances in Web programming techniques, such as WebAssembly (abbr. Wasm)~\cite{wasm} and WebGPU~\cite{Webgpu-w3c}, coupled with the ever-increasing processing power of edge devices.

However, current in-browser inference systems suffer from two major drawbacks. Firstly, their predefined kernels do not account for device diversity. Unlike cloud, edge devices are hetergenous, equipped with a range of CPUs, GPUs, memories, browsers, OS~\cite{steamSurvey}, and may also experience interference from other running applications. The \emph{one-for-all} approach delivers poor performance across devices. As we will show in Fig.~\ref{F.uncustomized_vs_customized_kernel_performance}, a predefined kernel can be several times slower than our device-customized kernels. Secondly, these systems lag behind cutting-edge Web programming techniques \ie WebGPU\cite{Webgpufordnn}. Their handwritten kernels for each new Web programming backend (\eg JavaScript, Wasm, WebGL\cite{Webgl}) necessitate significant rewriting efforts.

To address these challenges, kernel auto-generation techniques such as TVM\cite{chen2018tvm}, Ansor\cite{zheng2020ansor} and FlexTensor~\cite{flextensor} can be employed to automatically generate customized kernels without manual efforts. However, these techniques require \emph{ahead-of-time} kernel generation for known hardware.  Given a tensor computation, these techniques search for the potential optimal implementation from many possible implementations (e.g., different loop order). For each selected implementation candidate, they generate and compile the kernel code to evaluate on the target devices. This \textit{select-generate-evaluate} process will be repeated until the specified end condition is met.  To approach the optimal, this searching process can take hours even days to run, and also requires on-device kernel evaluation. This process is suitable for cloud with known and finite hardware. Unfortunately, Web applications are intentionally designed to function on diverse edge devices. Considering the huge number of different devices, generating kernels \emph{ahead-of-time} for each device is impractical. Consequently, achieving optimal in-browser inference performance for each edge device remains an unresolved challenge.

To tackle it, we rethink the specialties of Web. Compared to native precompiled inference systems, in-browser inference offers the distinct advantage of \emph{online kernel updating}. Furthermore, in-browser inference typically runs repeatedly over an duration, such as for video and document processing. This distinctive feature provides the opportunity and time budget for just-in-time (JIT) kernel customization after encountering the actual edge device.

Based on this insight, we present \sysname, the first in-browser DNN inference system with the unique ability to JIT auto-generate and continuously improve optimized kernels during inference for target edge devices, leading to a gradual speedup towards optimal performance.
Both CPU and GPU are supported through generating kernels in the state-of-the-art (SOTA) Web programming interfaces respectively, \ie Wasm for CPU and WebGPU for GPU. %

To realize this system, the main challenge lies in enabling JIT generation of optimized kernels, a feat that has never been accomplished before. This is due to the huge time cost of current optimized kernel generation, stemming from (1) the compiling cost and (2) the vast kernel optimization space. 
Tensor computations, implemented as multi-level loops, create a vast kernel optimization space due to variations in loop arrangements like tiling sizes. To find the optimized kernel, the kernel tuning process iteratively selects potential candidate from this space for compilation and evaluation on the target device. The compiling for each candidate can take minutes to complete numerous transforming passes. Previous works~\cite{liang2022romou, zhu2022roller, chameleon} try to reduce the space size or improve the searching method. However, the remaining space is still too large %
to enable JIT kernel optimization, or necessitates known hardware to build performance model ahead-of-time.

By comparison, \sysname can facilitate JIT generation of optimized kernels for diverse edge devices, based on our key findings of Web programming. %
(1) Web programming interface is designed with simple instruction sets and execution model for efficiency and security, which does not require complex compiling optimizations. Moreover, mostly compiling optimizations for Web programming interface are overlapped with kernel optimization space, \eg loop unrolling, rendering them unnecessary. (2) Strict Web requirements for security and portability convey consistent performance pattern across devices, \eg costly memory allocation. This consistency removes the need for related candidates in the kernel optimization space to be evaluated on target devices.

Based on the two findings, we propose two novel techniques accordingly. The first is \emph{Tensor-Web compiling co-design}. 
Taking Wasm compilation as an example. Rather than the separated tensor-level and language-level (\ie Wasm) compiling, \sysname employs a unified compiling pipeline that directly compiles tensor computation to in-browser executable, completely eliminating the costly invocation of separated language-level compiler \eg LLVM~\cite{LLVM} or Emscripten~\cite{emsscripten}. The unified compiling pipeline provides the capability of co-designing the tensor and language compiling optimizations to avoid redundant and ineffective ones. 
This new compiling pipeline dramatically reduces the cost per candidate, from minutes to milliseconds.

The second technique is \emph{Web-specific lite kernel optimization space}. The space is designed by offline consistent primitive setting decided by Web requirements, and online inconsistent primitive setting decided by target edge device.   
As Web requirements cause consistent performance patterns across devices, to identify their impact, we compose a microbenchmark suite that traverses the tensor compiling primitives (\ie code transformations conducted on tensor) such as loop order and unroll, in a \emph{one-variable-at-a-time} manner. The suite is evaluated offline to identify the efficient primitive settings. 
Besides the consistent ones, there are primitives \emph{inconsistent} across devices depending on hardware specifications. The dominant is tiling size that mostly impact the hardware utilization. A proper tile size can balance the contention between parallel hardware execution and advanced memory accesses. We use formulated kernel hardware usage and heuristics to select promising tile sizes to construct the lite kernel optimization space.
Consequently, the number of candidates in space is reduced, from millions to only dozens.

Based on the two techniques, we develop the \sysname. After the initial model and kernels are downloaded onto the target edge device, \sysname generates the lite kernel optimization space. Candidates in the space are compiled one-by-one using our unified compiling pipeline and evaluated on the device, interleaved with the inference process with limited overhead. Better kernels are continuously replaced online, gradually approaching the optimal. Considering the large number of clients on Web, candidate evaluation results and generated kernels are also crowdsourced from ones with similar device, achieving optimal kernels much faster.

We implement \sysname on both Wasm for CPUs and WebGPU for GPUs. It can run on devices with browsers installed that support Wasm or WebGPU.  Wasm is supported by mainstream browsers.
WebGPU, although still in its early stages,  shows great promise. Thanks to our JIT kernel auto-generation, \sysname is the first to supports WebGPU for complex models, serving as a strong showcase for our advantages.

\sysname is evaluated on representative modern models, with suitable size for edge devices, including T5~\cite{t5}, BART~\cite{bart}, GPT-2~\cite{gpt2}, RoBERTa~\cite{roberta} and Llama 2 7B~\cite{touvron2023llama}. Mainstream browsers are used \ie Chrome, Microsoft Edge, Firefox and Opera, which together take 87\% of market share~\cite{browsershare}. We evaluate on a range of smartphones, laptops and desktops, \eg Pixel 4, Vivo X30, SurfaceBook 3, Lenovo V9000, MagicBook and HP EliteDesk, equipped with ARM CPU (Cortex-A76, A78), Intel CPU (I9 12900H), AMD CPU (Ryzen 5800H), Intel GPU (HD 630), AMD GPU (Radeon), and Nvidia GPUs (RTX 3050, 3000, 3070Ti).
The results show within 30 seconds, \sysname can achieve up-to 20.1$\times$ faster kernels, and 8.2$\times$ faster model inference compared to SOTA inference frameworks. To summarize, our main contributions include: 
\begin{itemize}
    \item This paper proposes the first in-browser inference system that enables JIT optimized kernel generation.
    \item The Tensor-Web compiling co-design avoids the ineffective and redundant optimizations, reducing the compiling cost from minutes to milliseconds.
    \item The Web-specific lite kernel space design is guided by both Web programming requirement and efficient utilization of hardware resource, reducing the optimization space from millions to dozens.
    \item The evaluation is done on modern transformer models and a range of edge devices, achieving up to 8.2$\times$ speedup compared to SOTA frameworks.  
\end{itemize}

\section{Background and Motivation}

\subsection{DL Inference in Web Browsers}

Compared to cloud- or 5G edge-based solutions for DNN inference, on-device inference provides specific advantages in reducing cloud operation costs and providing improved privacy.  As reported by Microsoft and Google~\cite{zerocogs,google_edge}, on-device inference has the potential to achieve zero cost for providing DNN services to a large number of customers. Additionally, as stated in~\cite{cloudcyberattacks}, there is always a risk of data breaches when information travels over the internet. Therefore, even though network bandwidth is less of an issue these days, on-device inference attracts more and more attention compared to the cloud counterpart.

Nowadays, as many DNN models are directly integrated into Web applications~\cite{genai_ort}, in-browser inference on device in gaining momenta~\cite{microsoft2023ort, google2023tfjs,Webdnn,brainjs}. 
However, enabling DNN inference in modern Web browsers is nontrivial~\cite{Yunma19, wang2024exploring}. Due to the security considerations, the sandbox mechanism is widely used within browsers, which isolates Web applications, scripts, and other contents from %
the underlying system. 
The sandbox environment prevents malicious code from accessing and modifying system resources and settings, meanwhile it also restricts the usage of the sophisticated native DNN inference libraries, such as Eigen~\cite{eigen} for CPU and cuBLAS~\cite{cublas} for GPU. %

To make DL inference in browsers possible, alternative programming interfaces, hence backends, are proposed to use. JavaScript~\cite{javascript} is firstly leveraged to implement DL kernels and graphs in Web DL frameworks~\cite{google2023tfjs}. JavaScript has no-static data type and no vectorization support. Although some efforts like V8 Engine~\cite{v8} could significantly accelerate JavaScript code, the DL execution with it is still extremely inefficient in JavaScript environment.%

\begin{figure}[t]
  \centering
  \includegraphics[width=\linewidth]{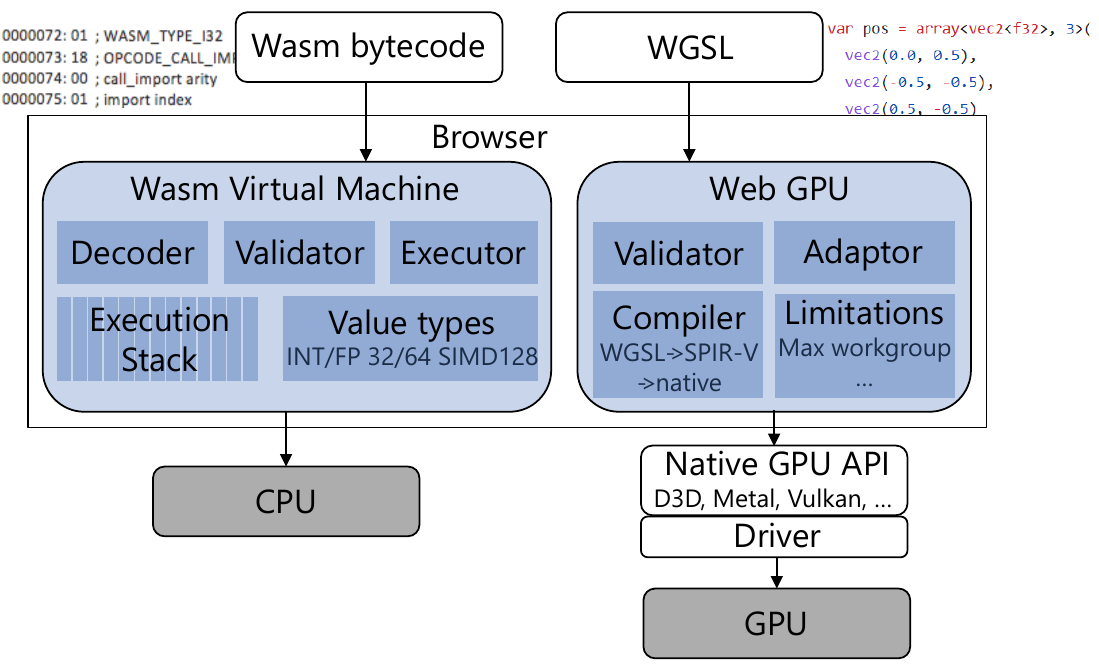}
  \vspace{-1em}
  \caption{The Wasm and WebGPU support in browser.}
  \Description{}
  \label{F.webbackend}
\end{figure} 

To cope with it, WebAssembly~(Wasm)~\cite{wasm} is considered. Wasm is a compact binary format.  Its runtime is a portable virtual machine running on the CPU. Fig.~\ref{F.webbackend} shows the Wasm implementation in browsers. %
Wasm code is delivered in low-level bytecode, which can be decoded and executed more efficiently in the virtual machine. The bytecode needs to be validated for security.  What's more, Wasm also takes advantage of advanced features of modern CPUs, \eg Single Instruction Multiple Data (SIMD). Therefore, it provides much better inference performance than JavaScript. Wasm is language-agnostic. High-level programming language like C and C++ could be compiled into Wasm bytecode.

GPUs are also utilized within browsers. For instance, WebGL has been integrated in TensorFlow.js, providing JavaScript interfaces to access GPU that originally enable rendering 3D graphics on Web pages. It is based on OpenGL ES 2.0~\cite{opengles}, a subset of OpenGL~\cite{opengl}. Thus, certain features are unavailable. Meanwhile as a rendering library, it failed to utilize computation pipelines in modern GPUs due to limited instructions.

To unleash the power of GPU, WebGPU, the successor of WebGL, is proposed. 
WebGPU provides
stronger computation ability, driving computation intensive DL kernels to execute more efficiently. WebGPU Shading Language (WGSL) is used to program. Fig.~\ref{F.webbackend} shows the implementation of WebGPU in browsers. While running in browser, the WebGPU kernel is translated to native GPU APIs, such as Vulkan~\cite{vulkan}. For portability, WebGPU also specifies limitations for the hardware usage. The validator is also introduced to check kernels for the security purpose.

Taking the advantages of the backends above, Web DL frameworks including TensorFlow.js~(TF.js) and Onnx Runtime Web~(Ort-Web), enable end-to-end in-browser inference for pretrained DL models. They all have relatively mature support for Wasm, and start to support WebGPU.  %
The kernels shipped within these frameworks are usually handwritten or ported from native DL frameworks, \eg TensorFlow~\cite{tensorflow}. To optimize the kernels, kernel auto-generators such as TVM~\cite{chen2018tvm} are extended for Wasm and even WebGPU.
However, generating kernels for Web usually takes extreme long time, \eg nearly 2 hours for one Matrix Multiplication (MatMul). Besides that, the performance of tuned kernels are almost far from the optimal, which we will discuss in the next. %

\vspace{-1em}
\subsection{Inference Performance Issues}
\label{subsec:performance_issues}

\begin{figure}[t]
  \centering
  \includegraphics[width=\linewidth]{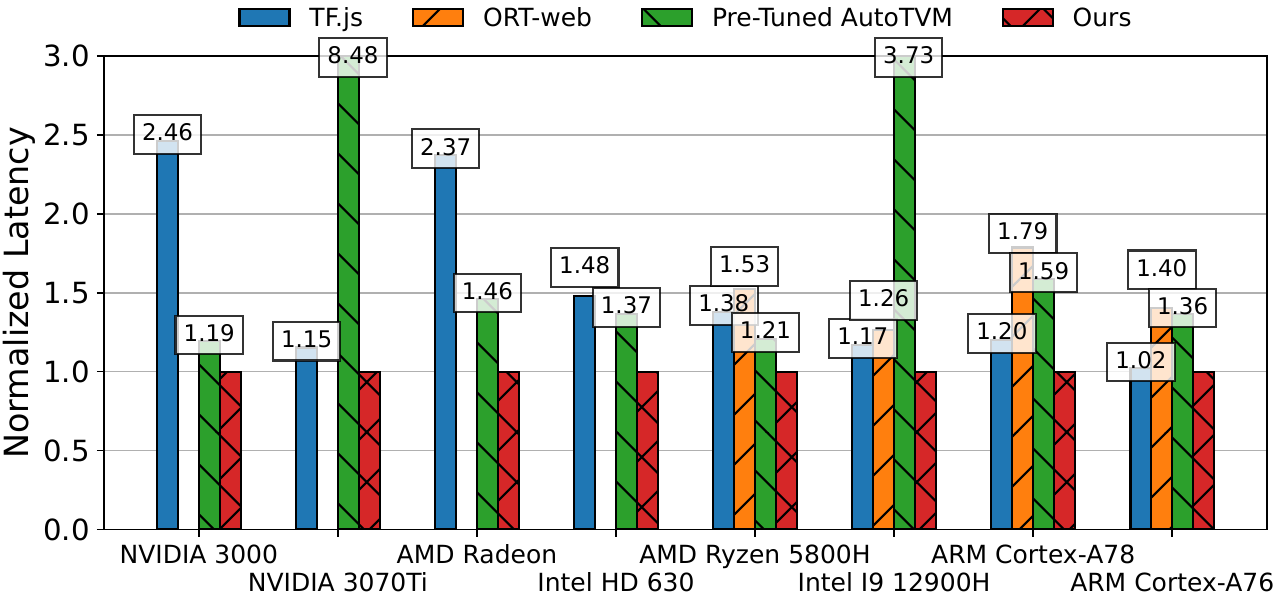}
  \vspace{-1em}
  \caption{The normalized kernel latency of handwritten, pre-tuned, and our \sysname for a MatMul  ([M,K,N]=[640,768,2304]).}
  \Description{}
  \label{F.uncustomized_vs_customized_kernel_performance}
\end{figure}

To understand in depth the DL inference performance in browsers, we conduct the preliminary study, using a MatMul kernel to demonstrate. %
We have the following observations:

\textbf{The one-for-all kernels are suboptimal across devices.} Web applications are running on millions of edge devices equipped with diverse hardware. Different hardware prefers different kernel implementations. However, instead of designing customized kernels for each type of devices, at present the SOTA in-browsers inference frameworks deliver kernels in a one-for-all style. For instance, TF.js and ORT-Web ship handwritten kernels on Wasm and WebGPU. We execute the one-for-all MatMul kernels from TF.js, ORT-web (only support Wasm), and pre-tuned AutoTVM (without tuning on the target device) on AMD 5800H desktop CPU, ARM Cortex-A78/A76 mobile CPUs, Nvidia 3000/3070Ti GPU and Intel 630 GPU. The inference latency is illustrated in Fig.~\ref{F.uncustomized_vs_customized_kernel_performance}. 

\begin{figure}[t]
  \centering
  \includegraphics[width=\linewidth]{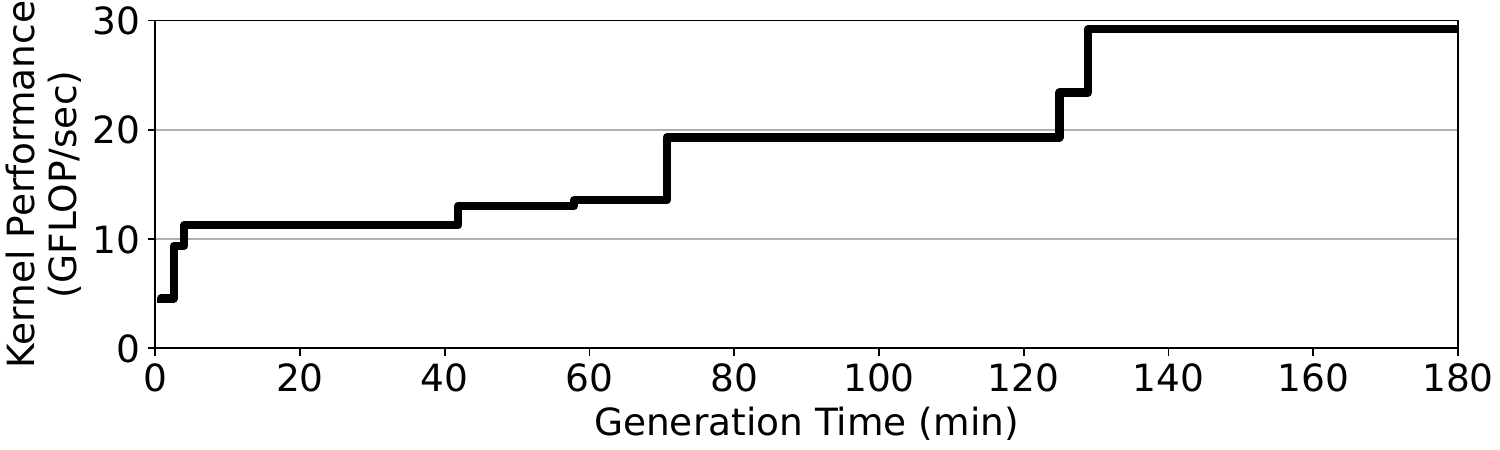}
  \caption{The generated MatMul kernel ([M,K,N]=[640,768,2304]) performance and generation time of TVM on AMD 5800H CPU.}%
  \vspace{-0.5em}
  \Description{}
  \label{F.tvm_generation_time_and_kernel_performance}
\end{figure}

The results indicate the performance of pre-defined kernels is suboptimal compared to our device-customized kernels. Moreover, a single pre-defined kernel exhibits a wide range of performance gaps on different devices. For instance, the kernel from TF.js demonstrates a slowdown ranging from as little as 2\% to as much as 246\% when compared to customized ones. Similarly, without tuning, the generated kernel from TVM shows a slowdown of 19\% to multiple times depending on devices. These results highlight the need for customized kernels tailored to each edge device.

\begin{figure}[t]
  \centering
  \includegraphics[width=1\linewidth]{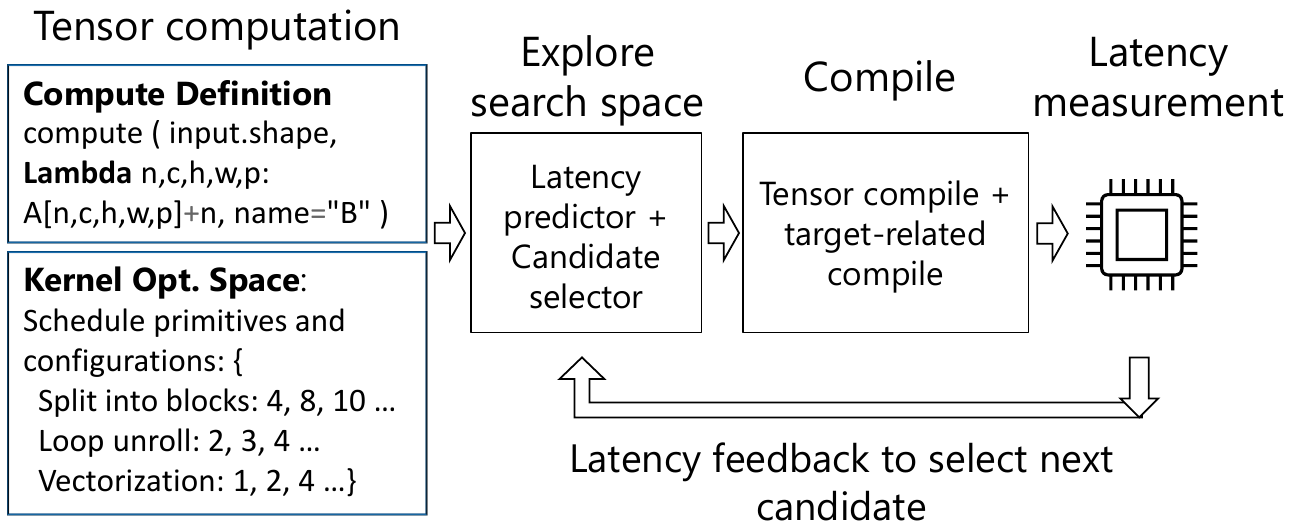}
  \caption{A common kernel generation pipeline.}%
  \vspace{-0.5em}
  \Description{}
  \label{F.tvm_compiling}
\end{figure}

\textbf{The one-for-each kernels are currently impractical in Web scenarios}. Based on the measurements presented above, one might consider generating kernels in a one-for-each style. However, this solution remains infeasible. We assessed the kernel generation time of TVM for a MatMul kernel on an AMD 5800H CPU device. It took nearly 2 hours to identify the kernel with the high performance (29.2 GFLOP/sec), with 437 tuning rounds. Typically, a deployed model contains several tens of kernels. Clearly, the one-for-each approach is impractical, particularly for Web scenarios where client diversity is substantial.

The prolonged kernel generation cost is due to two primary causes: the bloated compilation process and the exceedingly large kernel optimization space. 

Fig.~\ref{F.tvm_compiling} illustrates a common kernel generation pipeline. The tensor computation is defined in a domain specific language. Its potential kernel implementations, which composes a kernel optimization space, are defined by \emph{primitives} and the according configurations. A primitive is a kind of code transformation \eg loop unroll. A candidate from the kernel space can be described by a sequence of primitives and their configurations. The compiling process can then follow these primitives to conduct compiling IR (intermediate representation) transformations to generate the kernel. After that, the target language compiler \eg LLVM can be called to compile the kernel into executables for the target devices. 

As the combination blowup of loop arrangement, the kernel optimization space is huge. Our analysis shows the size of a naive space for a MatMul (384$\times$768$\times$768) in WebGPU is around 42\,M. Advanced searching algorithms and hardware performance models~\cite{liang2022romou,zhu2022roller,chameleon,flextensor,asymo} are normally employed to only select promising candidates for compilation and evaluation on the target device. Even so, thousands of candidates are generally evaluated before finding an optimized kernel implementation. The compiling cost for each candidate is around seconds to minutes depending on the kernel quality. The total generation cost of optimized kernel  will be hours.

Clearly, to reduce the kernel generation cost for JIT, we need to reduce the compiling cost for each candidate, and reduce the number of candidates in the space. Therefore, we propose \sysname. In the following sections, we will introduce the design principles and key techniques of \sysname.

\vspace{-1em}
\section{\sysname Overview}

\begin{figure}[t]
  \centering
  \includegraphics[width=\linewidth]{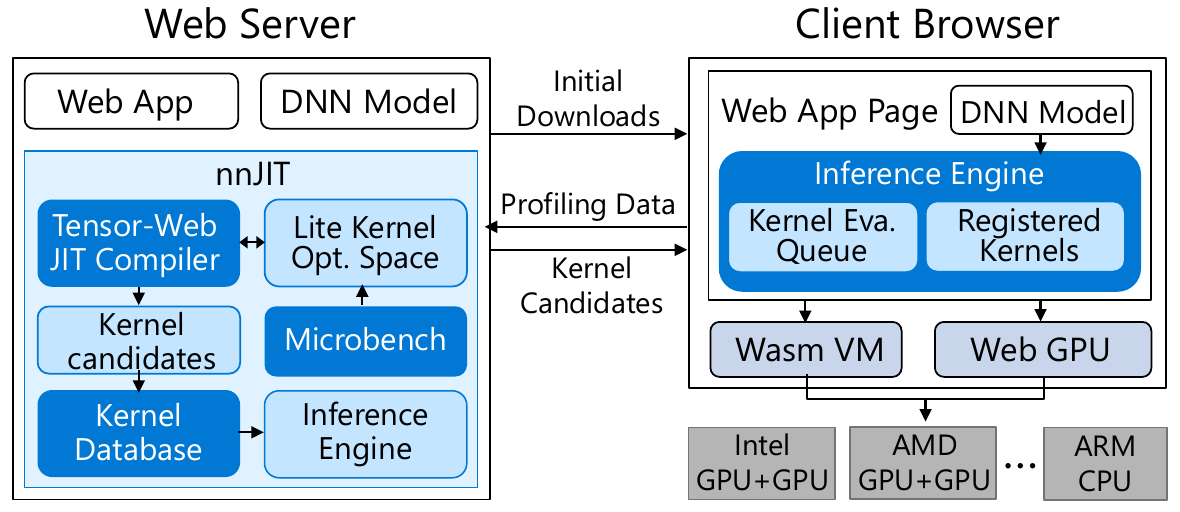}
  \vspace{-1em}
  \caption{Overview of \sysname.}
  \Description{}
  \label{F.overall_system}
\end{figure}

Fig.~\ref{F.overall_system} is the overview of \sysname. It consists of four modules: the \emph{tensor-web JIT compiler} for online kernel generation; the \emph{inference engine} for executing inference tasks in the browser; the \emph{micro benchmark suite} for offline exploration of the consistent primitive settings; and the \emph{kernel database} for storing optimized kernels tailored to known devices. The whole kernel generation and inference process facilitated by \sysname operates on both cloud and clients, as follows.

During the initialization phase, the browser on the client downloads the web page, the inference engine, the model and the initial kernels. The model encloses the weights and the optimized model graph (\eg operator fused) ready to deploy. The \emph{inference engine} parses the model graph, registers the kernel for each operator to execute, as well as manages the memory usage. The initial kernels are determined by the server, using the client device indicator, \eg device name and ids. If the hardware on client have been explored, the optimal kernels would be used from the \emph{kernel database} on server. Otherwise, the pre-defined and uncustomized ones are used meanwhile the JIT phase would be triggered. %

During the JIT phase, the \emph{tensor-web JIT compiler} on the server composes the lite kernel optimization space for each operator type. The compiler then subsequently generates the kernel for each candidate within the space. Between the server and the client, a kernel queue is established. Once a kernel is generated on server, it is pushed to the client via the queue. On client, the inference is executed repeatedly. Between every inference, the \emph{inference engine} retrieves one kernel from the queue and measures its latency. Based on the measurement, the newly retrieved kernel might be re-registered if it is significantly faster than the current registered one, ensuring that the more efficient kernel is utilized in subsequent inferences. After testing all the kernels in the queue, the best kernel along with the measurement results are reported to the server. The server would update the \emph{kernel database} according to the reports.

In accordance with our design, the \emph{tensor-web JIT compiler} of \sysname is lightweight and can be run either on the cloud or directly on clients. In our current implementation, we deploy it on the cloud, as this enables kernel reuse. Optimal kernels discovered by one device can be seamlessly shared with other devices possessing the same hardware through the cloud, thereby facilitating the concept of \emph{crowdsourcing}.

The online kernel generation combined with JIT-styled inference ensures optimal performance on Web across edge devices. To facilitate this, we propose two key techniques that significantly reduce the kernel generation cost, \eg from hours to milliseconds for a single kernel. In the following sections, we will introduce these techniques in detail.

\section{Streamline Compilation Pipeline through Tensor-Web Co-Design}

\begin{figure}[t]
  \centering
  \includegraphics[width=1\linewidth]{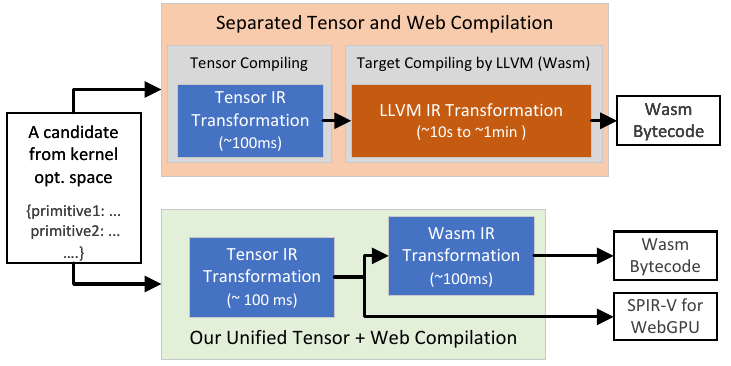}
  \caption{Our unified Tensor+Web compiling (lower half) compared to conventional separated Tensor and Web compiling (upper half).}%
  \label{F.tensor_compiler_on_web}
\end{figure}

Each candidate in the kernel optimization space needs to be compiled and evaluated on the client device. Current compiling takes minutes to complete. Even the space only has dozens of candidates, the total compiling will take hours, not possible to support the online optimized kernel generation. To reduce the cost, this section introduces the possibilities of removing target-related compilation (Sec.~\ref{subsec:remove}), mapping directly from tensor-level IR to Wasm IR (Sec.~\ref{sec:lowertensortowasm}), and only keep necessary optimization passes on Wasm IR (Sec.~\ref{sec:wasmopt}). Sec.~\ref{sec:compilingwebgpu} will briefly discuss compiling pipeline for WebGPU.   

\subsection{Unify Tensor-Web Compiling}
\label{subsec:remove}

\textbf{Costly target-related compilation.} As shown in Fig.~\ref{F.tensor_compiler_on_web}, the conventional compiling process of tensor computation consists of two main separated steps: the tensor-level compilation and the followed target-related compilation (\eg Wasm). Generally, they are designed separately by different communities, each with their own specific purpose. 

Tensor-level compilation transforms the tensor-level IR by the primitives and configurations of a picked candidate from the kernel optimization space, to generate a mapping of tensor computation to a loop arrangement. This process is independent of the target execution environment. Target-related compilation, on the other hand, aims to generate the efficient executables on the target environment from any high-level programs. Therefore, after tensor-level compilation generates the loop, a separate target-related compilation library such as LLVM is normally invoked to generate the executables. As these target-related compilation libraries aim to compile any general-purpose programs, there are many compiling passes, taking long time to complete.

Specifically, for Wasm, the target execution environment is Wasm virtual machine running within browsers. The target-related compilation library is LLVM or Emscripten. %
The compilation by LLVM/Em-scripten contributes the majority of total compilation cost, as shown in Fig.~\ref{F.tensor_compiler_on_web}.

\textbf{Feasibility of eliminating target-related compilation.}
We therefore explore the possibility to eliminate this target-related compilation, by identifying two opportunities.

Firstly, we could remove \emph{ineffective optimizations}. From the target perspective, Wasm is designed with a simple expression-based instruction set and a stack-based execution model~\cite{wasmspecs}, for the purpose of easy decoding, running efficiency, and security. Consequently, many sophisticated compiling optimizations would be not effective or necessary, thus not needed, such as the ones for register allocation, instruction reordering, and memory disambiguation. 

Secondly, we could remove \emph{duplicated optimizations}. From the tensor perspective, the kernel optimization space which includes numerous possible kernel implementations, also encompasses many of the target-related compilation optimizations. For example, the unrolled loop generated by LLVM optimization pass is very likely included in the kernel optimization space, which will be evaluated as well. The separated tensor-level and Wasm-level compilation cannot avoid the redundancy.     
In addition, the tensor computation defined domain specific languages need no complex compiling optimizations used by general-purpose programs, such as the dead code elimination. Thus, it could be further streamlined.

\textbf{Unified Tensor-Web compiling. }
The analysis above prompts us to redesign the compiling pipeline, which unifies the tensor-level and Wasm-level compilation as shown in Fig.~\ref{F.tensor_compiler_on_web}. It removes the separated target-related compiling invocation, and compiles tensor computation directly to the target executables \eg Wasm bytecode. The tensor-level IR is directly mapped to Wasm IR, and then mapped to Wasm bytecode. As a premise, Wasm is designed to be the compiling target of any high-level languages, including C and C++. It can also be the target of tensor-level IR.

The optimization passes of different level IR's are co-designed, retaining only the necessary and non-repetitive ones. Through analyzing the generated code performance, we find almost all the optimization passes in LLVM can be covered in kernel optimization space. Only the ones closely related to Wasm instruction definition will be additionally needed to apply on the Wasm IR as the figure shows. These passes are very light weighted, taking about 100\,ms to complete, tens or even hundreds of times less than calling LLVM.

\begin{figure}[t]
  \centering
  \includegraphics[width=\linewidth]{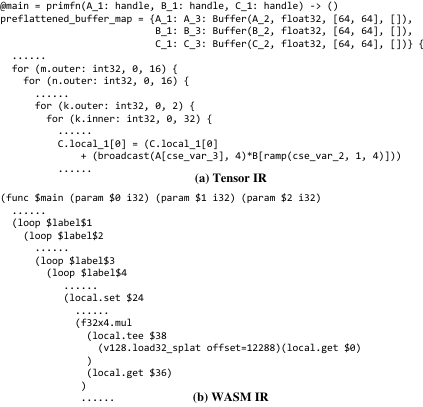}
  \caption{Lower tensor IR to Wasm IR for MatMul.}
  \label{F.condensed_compilation_pass_showcase}
\end{figure}

\vspace{-1em}
\subsection{Lower Tensor to Wasm}
\label{sec:lowertensortowasm}
The primary challenge in directly lowering tensor IR to Wasm IR involves determining how to effectively map the statement-based high-level tensor IR to the expression- and stack-based low-level Wasm IR. Fig.~\ref{F.condensed_compilation_pass_showcase} uses a code snippet to illustrate the differences between the two IRs, by using a MatMul implementation as an example. Wasm IR has only been lowered from LLVM IR before. LLVM IR is also a lower-level IR than tensor IR.  %
For example, LLVM IR has already lowered the high-level \texttt{for} statement in tensor IR.   

As Fig.~\ref{F.condensed_compilation_pass_showcase} shows, the tensor IR is represented as a sequence of statements, such as the \texttt{for} loop statement. %
Wasm IR, on the other hand, is composed of a sequence of expressions (enclosed by the parenthesis in the figure). Each expression is evaluated to produce a value. Wasm virtual machine to run Wasm bytecode is stack-based, in which instructions manipulate an implicit operand stack, popping argument values and pushing result values. This design is to fit the sandboxed and resource-limited environment of browsers. 

The lowering of tensor IR to Wasm IR needs to consider both the expression and stack execution order.   

\textbf{Map~\texttt{for} statement to an expression block.} %
Algorithm \ref{A.tensor_ir_to_wasm_ir_for_loop} shows the transform of the \texttt{for} statement.
We construct a nested sequence of expressions as a block enclosed by the Wasm \texttt{loop\&end} instructions for this statement. 

As shown in Algorithm \ref{A.tensor_ir_to_wasm_ir_for_loop}, the sub-expressions, \eg  loop variable calculation, are created while traversing the \texttt{for} node of the tensor IR (line 2-7). Then the expressions will be nested together as the execution order of the stack (line 8-10). 
During execution, the \texttt{br\_if} will pop the condition result from the stack, and decide whether to branch to the loop label (line 5). 
The \texttt{loop} instruction introduces an implicit label, which serves as the target of the branch instruction. During the actual stack execution, the \texttt{loop} instruction pushes a new entry onto the control stack, and record the stack height. If the branch is taken, the stack pops up to the block's height before and proceed to the end of the block.

\begin{algorithm}[t]
  \footnotesize
  \SetKwInOut{Input}{input}\SetKwInOut{Output}{output}
  \Input{ForNode of Tensor IR $forNode$}
  \Output{LoopExpression of Wasm IR $loopExpr$}
  \Comment{{\scriptsize for(loopVar=begin;loopVar<end;loopVar+=stride) body\;}}
  \SetKwProg{Fn}{Func}{:}{}
  $loopVar$ $\leftarrow$ createWasmVar()\;
  $initLoopVarExpr$ $\leftarrow$ makeLocalSet($loopVar$, $forNode$.begin)\;
  $ltExpr$ $\leftarrow$ makeBinary(Op::Lt, $loopVar$, $forNode$.end)\;
  $brIfExpr$ $\leftarrow$ makeBreak($loopVar$.label, $ltExpr$)\;
  $addLoopVarExpr$ $\leftarrow$ makeBinary(Op::Add, $loopVar$, $forNode$.stride)\;
  $bodyExpr$ $\leftarrow$ VisitStmt($forNode$.body)\;
  $innerExpr$ $\leftarrow$ makeBlock($bodyExpr$, $AddLoopVarExpr$, $brIfExpr$)\;
  $loopExpr$ $\leftarrow$ makeLoop($loopVar$.label, $innerExpr$)\;
  $loopExpr$ $\leftarrow$ makeBlock($initLoopVarExpr$, $loopExpr$)\;
\caption{Lower Tensor IR to Wasm IR for loop}
\label{A.tensor_ir_to_wasm_ir_for_loop}
\end{algorithm}

\begin{figure}[t]
  \centering
  \includegraphics[width=\linewidth]{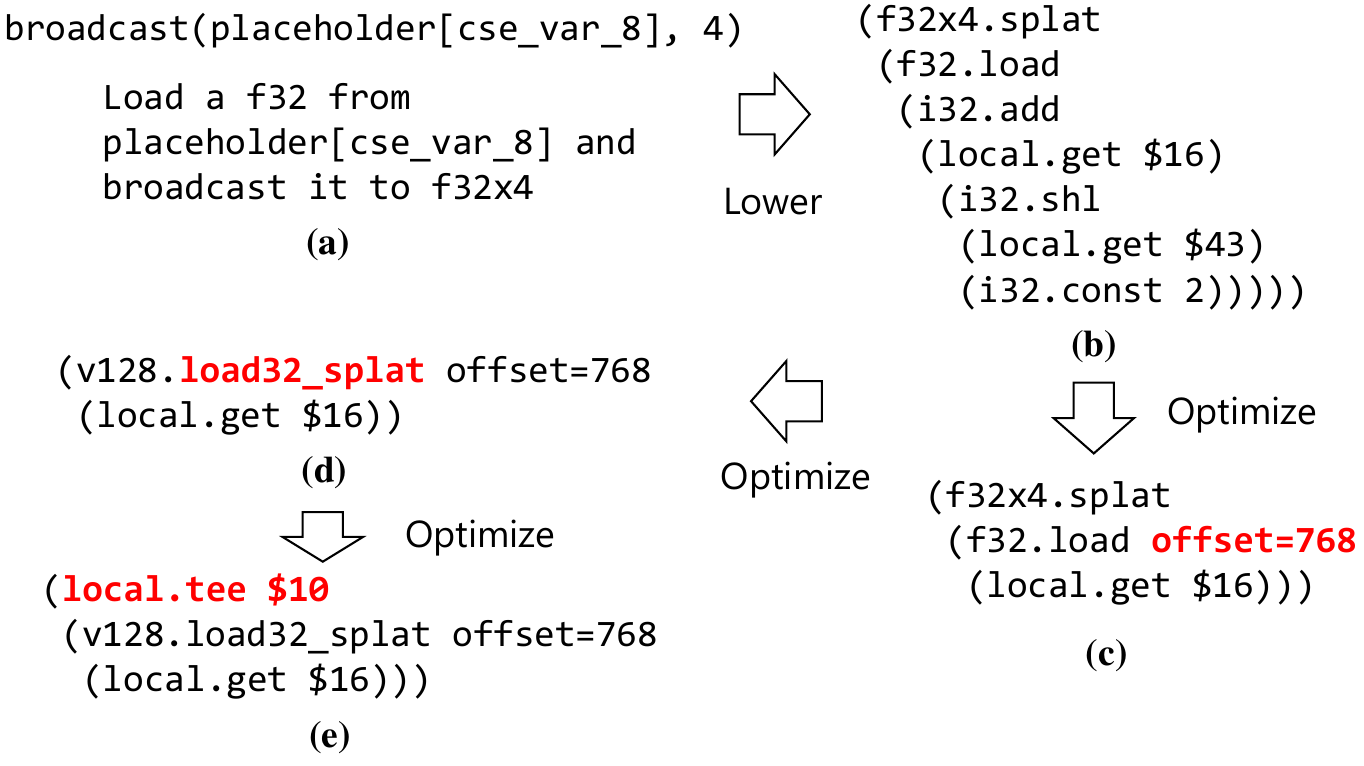}
  \caption{Wasm IR transformation by applying each of compiling passes: (a) tensor IR for a broadcast load statement; (b)lowered Wasm IR; (c) Wasm IR after offset load/store pass; (d) Wasm IR after combined instruction pass; (e) Wasm IR after load/store to variable pass.   
  }
  \Description{}
  \label{F.tir_to_wasm_ir_optimized_passes}
\end{figure}

\subsection{Compiling Optimizations for Wasm IR}
\label{sec:wasmopt}
As stated above, our compiling pipeline applies the optimization passes related to Wasm instruction definition to the Wasm IR.  
Only three passes are needed, as shown in Fig.~\ref{F.tir_to_wasm_ir_optimized_passes}: 1) offset load/store, 2) load/store to variable, and 3) combined instruction. Each pass is explained in detail below. 

(1) Offset load/store pass is to eliminate constant address calculation for load/store instructions. 
Wasm code execution accesses a linear memory in the Wasm virtual machine. Wasm provides the offset augmented load/store instruction to avoid the address calculation. This pass is to utilize this instruction. By applying it as shown in Fig.~\ref{F.tir_to_wasm_ir_optimized_passes} (b, c), five additional instructions  can be eliminated for each load/store. This optimization can speed up generated kernels by 2.7$\times$.

(2) Combined instruction pass is to eliminate separated instructions if possible. It is to apply the combined instruction, \ie \texttt{v128. load\_splat}, provided by Wasm. This \texttt{load\_splat} combines \texttt{load} and \texttt{splat} instructions into one that loads a single lane and duplicate it to all lanes of the vector.  

(3) Load/store to variable pass is to eliminate repeated stack popping and pushing. 
Wasm \texttt{local.tee} instruction duplicates the top of the stack to a variable for later use. %
This pass applies this instruction to replace repeated load/store of the same memory address, and avoid the repeated stack pushing and popping of this variable. This can reduce kernel latency by $\sim$7\%. 

Just applying these light-weight optimization passes, the Wasm byte codes generated by \sysname has no noticeable performance or byte code difference. The compiling latency can be accelerated by 
up to 125$\times$ compared to SOTA practice.
\subsection{Compiling for WebGPU}
\label{sec:compilingwebgpu}
In contrast to Wasm, which is executed in the browser's virtual machine, WebGPU implementation in browser essentially only translates WebGPU APIs into native GPU APIs (with limited optimization passes), while the target-related compilation is handled by the GPU driver. As illustrated in Fig.~\ref{F.tensor_compiler_on_web}, it is not possible to eliminate the separated invocation of this target-related compilation; instead, we can only lower tensor IR to SPIR-V\cite{spirv}, a portable IR supported by both native GPU APIs and WebGPU. The reduction in compilation time for WebGPU is achieved through the kernel optimization space design, which will be discussed in Sec.~\ref{sec:kernel_search}.

\section{Accelerate Kernel Tuning with Web-Specific Lite Space}
\label{sec:kernel_search}

To reduce the vast kernel optimization space, we propose the web-specific lite kernel space design based on two guidelines: the web specific requirements (Sec.~\ref{subsec:web_sapce}), and the efficient utilization of hardware resources (Sec.~\ref{subsec:device_space}). Existing works%
also aim to shrink kernel optimization space for inference on native hardware, but these spaces are either still too large to be evaluated online or require pre-defined hardware performance models. We will show that for in-browser inference, considering the two guidelines leads to a lite space size of just dozens, which can be evaluated online. Moreover, numerous edge devices offer the unique opportunity for crowdsourcing global optimal kernels (Sec.~\ref{subsec:crowdsourcing}).

\begin{figure}[t]
  \centering
    \begin{subfigure}[c]{0.24\textwidth}
      \includegraphics[width=\textwidth]{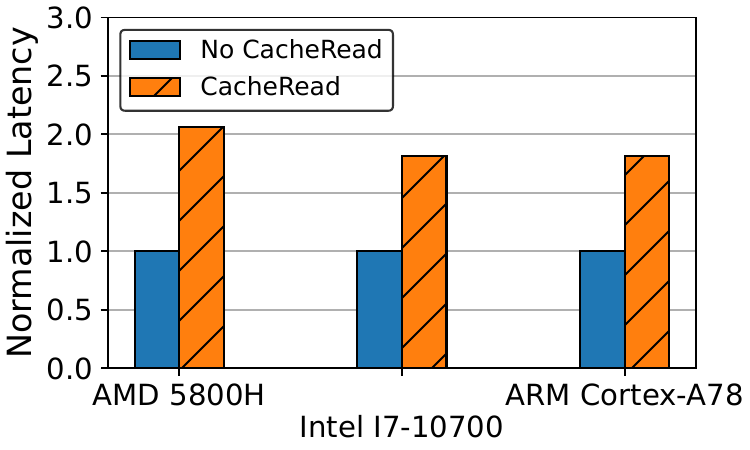}
      \caption{WASM}
      \label{F.wasm_schedule_cache_read_and_kernel_latency}
    \end{subfigure}
    \begin{subfigure}[c]{0.23\textwidth}
      \includegraphics[width=\textwidth]{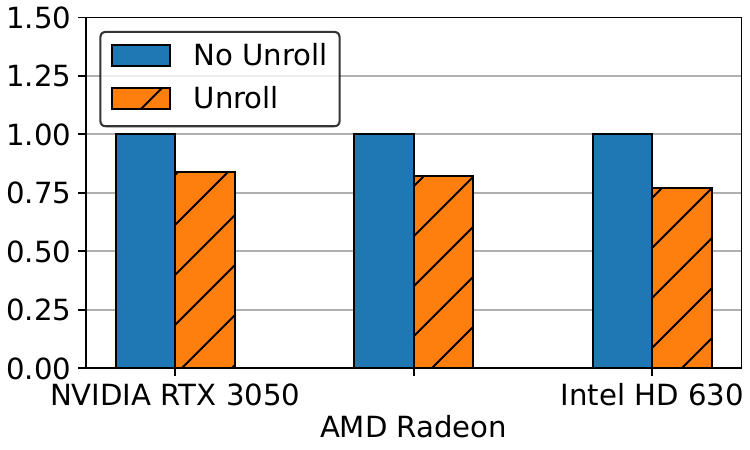}
      \caption{WebGPU}
      \label{F.webgpu_schedule_unroll_and_kernel_latency}
    \end{subfigure}
  \caption{The MatMul kernel ([M,K,N]=[640,768,2304]) latency comparison of different primitive settings. The advanced setting is consistent across devices.}
  \vspace{-1em}
  \Description{}
  \label{F.wasm_schedule_cache_read_write_and_kernel_latency}
\end{figure}

\vspace{-1em}
\subsection{Web-guided Offline Space Reduction}
\label{subsec:web_sapce}
 Web programming is aimed at achieving portability and security. For instance, both Wasm and WebGPU implement rigorous validation processes to prevent malicious or erroneous code, such as type errors, memory overflow, out-of-bounds access, and invalid jumps. These specialties convey consistent kernel performance patterns across devices. The related kernel implementations do not need to be evaluated on every device, which can significantly reduce the number of candidates within the kernel optimization space.

\textbf{Performance pattern of Web programming}. To illustrate the performance impact, Fig.~\ref{F.wasm_schedule_cache_read_write_and_kernel_latency} compares a MatMul latency with different primitive settings, \ie  \texttt{cache\_read} on and off for Wasm, and the \texttt{unroll} on and off for WebGPU as examples. The performance shows the same pattern across devices. Disabled \texttt{cache\_read} and enabled \texttt{unroll} always achieve better performance. What is more, they are also against the common setting for native kernels. The reasons are explained as follows. %

The \texttt{cache\_read} primitive creates a small buffer that can reside in different memory levels. As a nested loop in a kernel is mapped to various levels of tiling on the hardware. The small buffer can load a tile to improve data locality. For native kernel execution, the \texttt{cache\_read} does improve performance on many devices. However, when it comes to Wasm kernels, the performance is reduced on all tested devices as shown in Fig.~\ref{F.wasm_schedule_cache_read_write_and_kernel_latency}. The performance decrease is attributed to the costly Wasm validation process for memory allocation.

The \texttt{unroll} primitive explicitly unrolls the loop to reduce the loop related overheads. In native inference, the \texttt{unroll} primitive does not impact kernel performance on many devices, as the native GPU compiler can conduct loop unrolling optimization as needed. However, WebGPU only triggers a weak level of compiling optimization in native GPU to facilitate the quick response of web applications. As a result, the \texttt{unroll} primitive needs to be specifically set for tensor compiling to achieve better performance.

\textbf{Discovery of Web-consistent primitive settings}. Although we have demonstrated two typical examples of primitive settings, it remains challenging to discover all such primitives with cross-device consistent settings. To minimize human efforts, we propose developing a microbenchmark to automatically detect these primitives. The benchmarking is a one-time effort, as it is only related to the Web techniques used for backends, such as Wasm and WebGPU.

The microbenchmark suite automatically traverses all the primitives for a common-sized MatMul kernel (specifically with a shape of 4K$\times$4K$\times$4K in practice). The \emph{one variable at a time} method is used to change the setting of only one primitive at a time, such as {cache\_read on/off}. The suite is evaluated offline on multiple testing edge devices. We then compare the measurements across these testing devices. If the results are consistent, we set the primitives accordingly, \eg\texttt{cache\_read} off and \texttt{unroll} on. Consequently, we can fix these settings when constructing the kernel optimization space, hence reducing the space. If the results are inconsistent, we consider them as device-dependent primitives. These will be processed in the device-guided online space construction module, allowing for adjustments based on specific device characteristics to optimize performance.

\begin{figure}[t]
  \centering
  \includegraphics[width=0.6\linewidth]{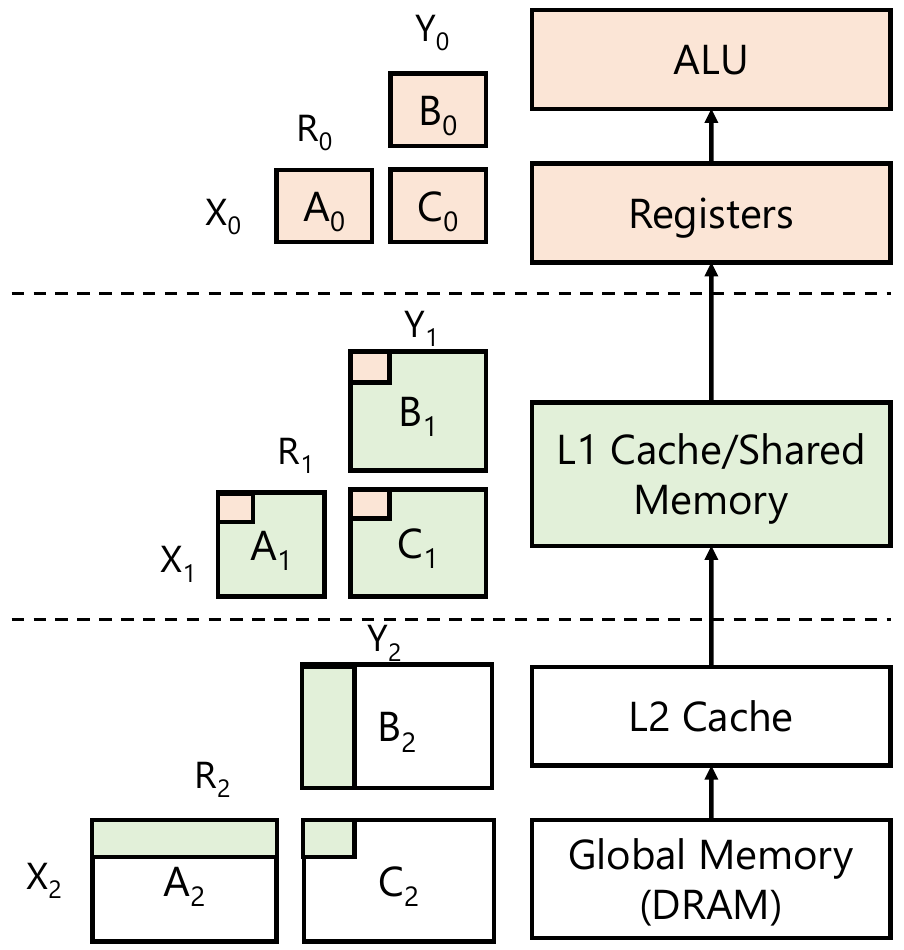}
  \caption{Tiles on the memory hierarchy for a MatMul.}
  \Description{}
  \label{F.tile_computing_and_memory_hierarchy}
\end{figure}

\subsection{Device-guided Lite Space Building}
\label{subsec:device_space}

Microbenchmark results remove the device-consistent settings from the kernel optimization space. The left ones are inconsistent across devices. This space is still large in the size of tens of thousands. This section will use formulated kernel hardware usage and heuristics to build the lite space with promising candidates for JIT evaluation on target devices.

\textbf{Rational for heuristics and formulation. }  
Tensors are divided to tiles for parallel execution. As shown in Fig.~\ref{F.tile_computing_and_memory_hierarchy}, the innermost loop tile is loaded to the registers by a thread.  The second level tile is loaded to the L1 cache/shared memory by a block. For GPU programming, a block is a group of threads that are executed together on a core.  The tile size prominently impacts the hardware utilization of a kernel implementation, and thus the kernel performance. 

Fundamentally, a tile size with efficient hardware utilization is to balance (1) the use of parallel computation units for fast computation and (2) the advanced memory storage \eg registers for fast data accesses. However, the two are normally conflicted with each other. More blocks and threads on a core can better saturate the parallel computing units and hide memory access stalls. However, since threads on a core share the registers and cache, too many threads may overuse the resources. On the other hand, fewer threads within the register and cache limit would under-use the parallel units. The sweet spot balancing the two highly depends on exact hardware, including the size of advanced memories, the computation and memory bandwidth, and the quality of compiling. It has to be evaluated on device to obtain.

\begin{table}[t]
  \footnotesize
  \caption{Formulation of tile-based kernel hardware utilization and heuristics of web-specific lite space.} 
  \vspace{-1em}
  \label{T.specific_paramter}
  \begin{tabular}{|c|c|}
    \hline
    \multicolumn{2}{|c|}{\makecell[l]{\textbf{Params.:} (Symbols follow Fig.~\ref{F.tile_computing_and_memory_hierarchy})\\
    $x_0$, $r_0$, $y_0$: size of the inner-most tile loaded by a thread, \\
    $x_1$, $r_1$, $y_1$: size of the second tile loaded by a block, \\
    $Reg_{thread}$: number of registers used by a thread,  \\
    $L1_{block}$: size of L1Cache or shared mem. used by a block, \\
    $Thread_{block}$: number of threads used by a block, \\
    $Warp_{block}$: number of warps used by a block. \\
    $Block_{core}$: number of blocks sharing a core
    }}\\
    \hline

    \multicolumn{2}{|c|}{\makecell[l]{\textbf{Hardware resource usage:} \\
    $Reg_{thread}$= $(x_0 \cdot r_0) + (r_0 \cdot y_0) + (x_0 \cdot y_0)$, \\
    $L1_{block}$ = $(x_1 \cdot r_1) + (r_1 \cdot y_1) + (x_1 \cdot y_1)$, \\
    \textbf{For WebGPU}:  \\
    $Thread_{block}$ = $(x_1 / x_0) \cdot (y_1 / y_0)$, \\
    $Warp_{block}$ = $Thread_{block} / $\sc{WARP\_WIDTH}, \\
    $Block_{core}$ = $min($RegLimit, L1Limit, WarpLimit$)$, \\
         $\begin{aligned}
           &\hspace{1.5cm} RegLimit = \frac{TOTAL\_REGS\_PER\_CORE}{Reg_{thread} \cdot Thread_{block}}, \\
           &\hspace{1.5cm} L1Limit =  \frac{L1\_SIZE\_PER\_CORE}{L1_{block}}, \\
           &\hspace{1.5cm} WarpLimit = \frac{TOTAL\_WARPS\_PER\_CORE}{Warp_{block}} \\
     \end{aligned}$ \\
     \textbf{For WASM}: $Thread_{block}=1, Block_{core}=1$
    }}\\
    \hline

    \makecell[c]{Heuristics\\ for all} & \makecell[l]{
    \\[-1em]
    1. $SIMD\_width \leq 128$ \\
    2. $x_0, r_0, y_0, x_1, r_1, y_1$: power of 2}\\
    \hline

    \makecell[c]{Heuristics\\ for WASM} & \makecell[l]{
    \\[-1em]
    3. $L1_{block} \leq L1\_SIZE\_PER\_CORE$ \\
    4. $Reg_{thread}$ = TOTAL\_REGS\_PER\_CORE}\\ 
    \hline

    \makecell[c]{Heuristics\\ for WebGPU} & \makecell[l]{
    \\[-1em]
    5. $Thread_{block} \leq 256$ \\
    6. $L1_{block} \leq 16\,KB$ \\
    7. $Block_{core} = RegLimit$} \\
    \hline
  \end{tabular}
\end{table}

\textbf{Heuristics for efficient hardware utilization. }   
We therefore formulate the hardware utilization based on the tile sizes, as shown in Table~\ref{T.specific_paramter}. These formulas require only the identification of the edge device type to ascertain hardware constraints, such as cache and register size. This device type could be obtained by Web programming interface. No other prior knowledge about devices are needed. By calculating the hardware utilization of each candidate and filtering the candidates by the heuristics, \sysname can build the lite kernel optimization space for the hardware.

Taking the lite optimization space construction for WebGPU as an example,
the tile size determines the utilization of registers, L1 cache, threads, and warps of a block. Given the total available registers, L1 cache, and warps of a GPU core, we can calculate the number of blocks that can share a GPU core ($Block_{core}$). To fully utilize the registers, Heuristic 7 guides the lite space to only include candidates whose number of blocks constrained by total registers to avoid wasting this fastest storage.  Besides the hardware constraints heuristics (2, 3, 4 and 7), the Wasm and WebGPU specifications based heuristics (1, 5 and 6) are also considered.

The candidates that satisfy the heuristics will be in the optimization space. For online evaluation on devices, the candidates in the space will be evaluated in the order from the ones with max blocks per core to the ones with min blocks per core. Note in the actual calculation, there is a relaxation ratio for the hardware resources, since other variables in a kernel implementation also use registers or caches.

Finally, by applying both the Web-guided space reduction and the device-guided space building, 
our web-specific lite kernel optimization space only includes a few dozens of candidates, six orders of magnitude smaller than the naive search space, which is able to be evaluated online. %

\textbf{Potential kernel optimization for memory.} Our current space design prioritizes latency considerations. We could also incorporate memory access or memory usage constraints into the heuristics for web-specific lite space. For instance, the search object could be the kernel implementation with the fewest memory accesses. Alternatively, we could constrain the value of  $L1_{block}$ in Table \ref{T.specific_paramter}, to reduce the usage of shared memory for WebGPU.  Exploring extensions of \sysname to other metrics represents a potential avenue for future work.

\vspace{-1em}
\subsection{Crowdsourcing and Kernel Zoo}
\label{subsec:crowdsourcing}

We have constructed a lite kernel optimization space. %
During deployment, we recognize a potential issue arising from the diverse nature of deployment environments, including various background workloads and hardware utilization levels. This may cause variance in assessed latency, potentially affecting our choice of the optimal kernel. To mitigate these concerns, we propose an \emph{extended} kernel space.

\textbf{Extended kernel space.} For each device, we enhance the lightweight kernel space using an exploration-and-exploitation approach. The extended kernel space typically comprises two sets of candidates: 1) the exploration set, which includes the original lightweight kernel set and may be empty if optimal candidates for the device have already been discovered; 2) the exploitation set, obtained from the \emph{crowdsourcing} module, which gathers and sends optimal kernel candidates to new devices with similar hardware specifications for further validation. Overall, the number of extended candidates is approximately one-tenth of the lite kernel space.

\textbf{Crowdsourcing and the kernel dataset.} The diverse nature of web clients provides us with the opportunity to engage in \emph{crowdsourcing}. The fundamental concept is that the searched optimal kernel implementations can be shared among devices with identical hardware. To facilitate this, we employ two designs: (1) we leverage the hardware ids as well as profiled hardware primitives as a criterion to ascertain whether devices can share the same generated optimal kernel. In particular, we form the primitive vector as $\vec{\rho}=\langle\rho_{i}\rangle, \rho_{i}\in\{0, 1\}$, where $\rho_{i}$ denotes the $i$ th primitive obtained from the micro benchmark. (2) In order to identify the best generated kernel, we adopt a majority voting strategy. Clients submit top-N (5 in our implementation) fastest implementations for a given kernel with the ranked weights.
We also introduce the kernel dataset. We take as the primary index key $(t, s, \vec{\rho}, id)$, where $t$ is the kernel type, $s$ denotes the kernel shape, $\vec{\rho}$ is the primitive vector and $id$ is the device id. Although the background noise is inevitable for latency measurements on edge devices, the majority voting result could reflect the likely kernel latency under the typical background workload.

\vspace{-1em}
\section{Implementation}

The Tensor-Web co-designed compilation pipeline in \sysname is built upon TVM \cite{chen2018tvm}. Particularly, for Wasm, we introduce a new compilation target (\ie Wasm IR) in TVM, by leveraging Binaryen \cite{binaryen} Wasm IR constructor.  We develop the \code{WasmModuleNode} to enable lowering tensor intermediate representation (IR) to Wasm IR. We implement two crucial functions, \code{wasm::Builder} and \code{wasm:: ModuleWriter}, to construct Wasm IR and compile Wasm binary. For WebGPU, we use the native GPU driver to compile.

\begin{table}[t]
\caption{The web-specific lite space for a MatMul kernel (M=K=N=4096) on WebGPU.}
\vspace{-1em}
\footnotesize
\begin{tabular}{|l|l|}
\hline
\textbf{Primitives} & \textbf{Configures (symbols follow Table~\ref{T.specific_paramter})}  \\ \hline
\makecell[l]{Cache Read \\ Cache Write \\ Reorder \\  Bind \\ \\ Unroll \\ Vectorize \\ Tile Size \\ } & \makecell[l]{[Yes,No] \\ Yes \\ $r_2,y_2,x_2,r_1,y_1,x_1,r_0,y_0,x_0$ \\ $y_2$ $\rightarrow$ block.y, $x_2$ $\rightarrow$ block.x \\ $y_1$ $\rightarrow$ thread.y, $x_1$ $\rightarrow$ thread.x \\ $y_0$ \\ $x_0$ \\ $(r_0,y_0,x_0)\in[4, ..., 32]$;  $(r_1,y_1,x_1) \in [32, ..., 256]$} \\ \hline
\end{tabular}
\label{T.web_specific_lite_space}
\end{table}

To create the lite kernel space for \sysname, we extend TVM by incorporating web-specific scheduling templates. In these templates, we set the configurations for web-consistent primitives and define the search space for device-dependent primitive configurations selected by heuristics. Table~\ref{T.web_specific_lite_space} shows an example lite kernel optimization space we build for a MatMul. We implement the microbenchmark via the evaluation MatMul kernel with the shape of 4\,K$\times$4\,K$\times$4\,K, which are then compiled by the tool chain described above.
We also adapt the in-browser inference runtime based from TVM, which is compatible to models in TF and ORT format.

To deploy \sysname, we piggyback existing in-browser DNN deployment flow. Specifically, we introduce the kernel generation pipeline to the existing kernel distribution pipeline on the cloud side and incorporate kernel evaluation and kernel replacement function during inference on the edge side, Moreover, the cross-platform nature of web applications makes this integration a one-time effort for all devices. Overall, \sysname comprises 2085 new lines of Python code, 1671 new lines of C++ code, and 564 new lines of JavaScript.

\begin{figure*}[!th]
    \centering
    \begin{subfigure}[t]{0.55\textwidth}  
        \centering 
        \includegraphics[width=\linewidth]{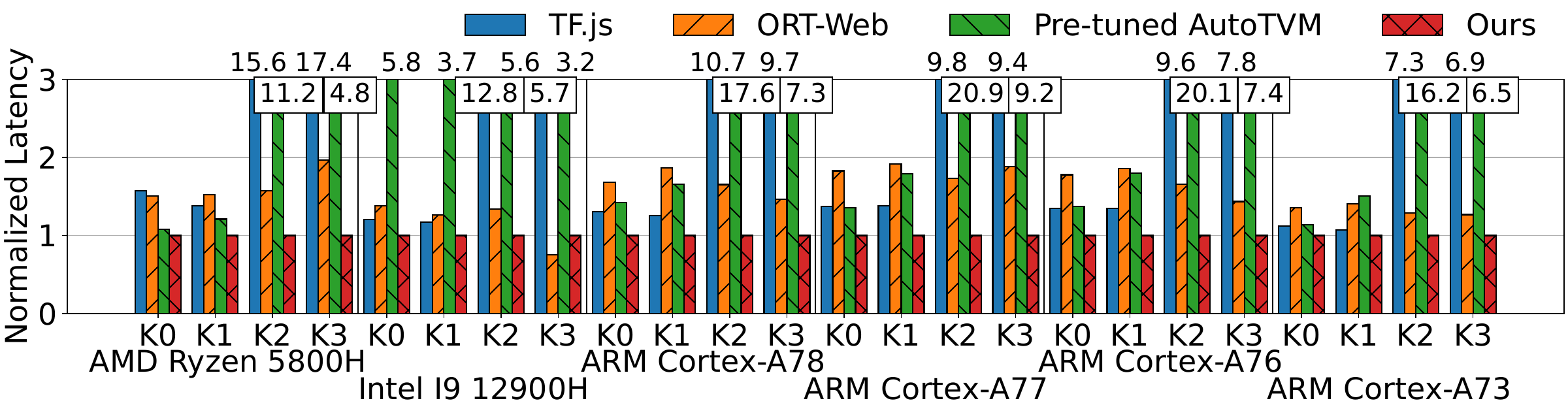}
        \caption{WASM}
    \end{subfigure}
    \begin{subfigure}[t]{0.42\textwidth}
        \centering
        \includegraphics[width=\linewidth]{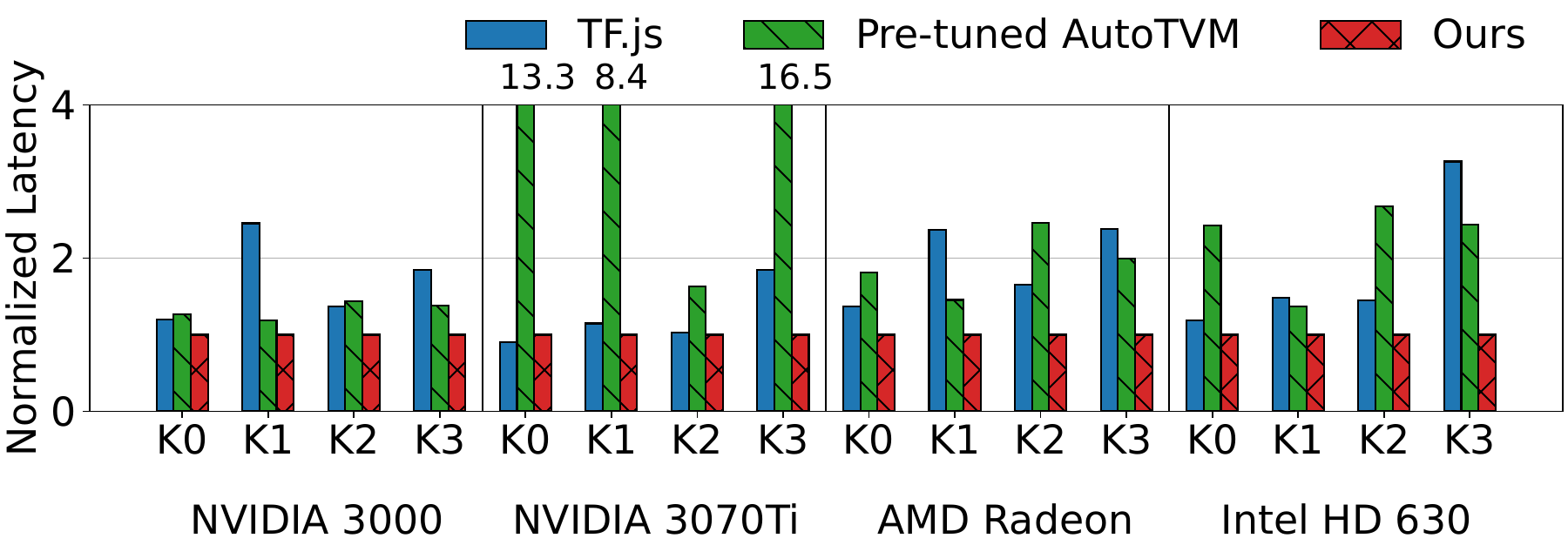}
        \caption{WebGPU}
    \end{subfigure}
    \caption{Kernel latency executed with TF.js, ORT-web, pre-tuned AutoTVM as well as \sysname on Chrome.} 
    \label{F.kernel_latency_comparison}
\end{figure*}

\begin{figure}[!th]
    \centering
    \begin{subfigure}[t]{0.48\textwidth}
        \centering
        \includegraphics[width=0.95\linewidth]{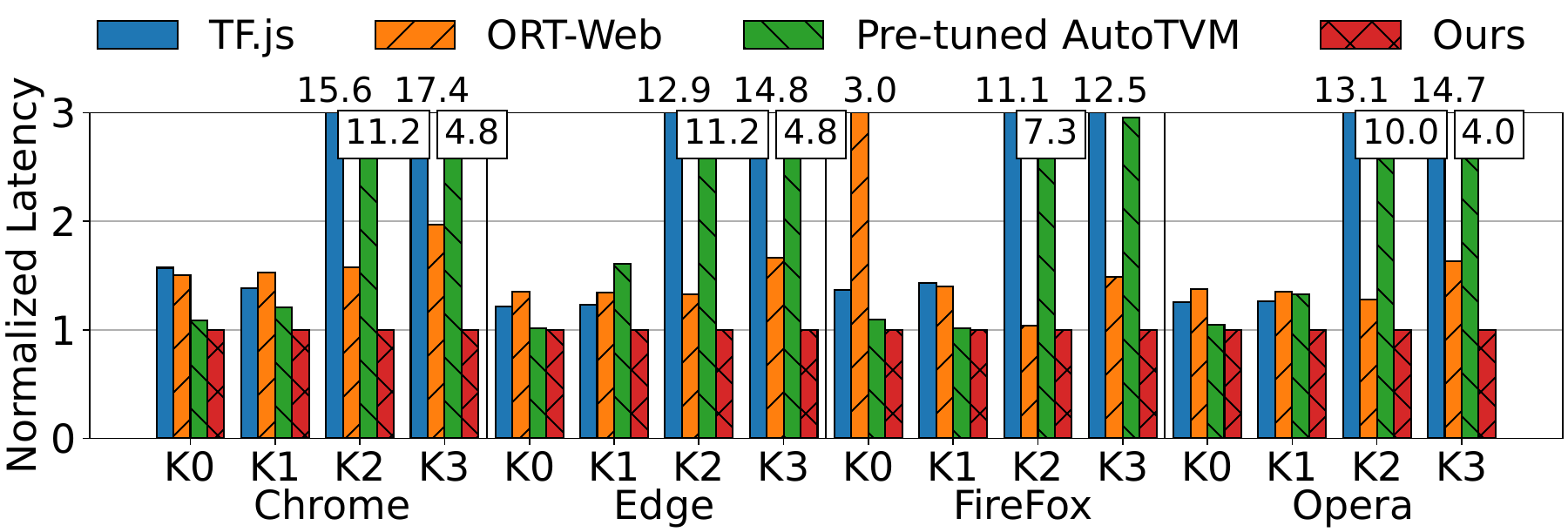}
    \end{subfigure}
    \vspace{-1em}
    \caption{Kernel latency on four browsers with AMD Ryzen 5800H for WASM with \sysname as well as baselines. }
    \label{F.kernel_latency_comparison_on_browsers}
\end{figure}

\section{Evaluation}
\subsection{Experiment Setup}

\textbf{Hardware and browsers.} We conduct experiments on smartphones, laptops and desktops, a total of six devices, including Pixel 4, Vivo X30, Mate 20, Honor 70, SurfaceBook 3, Lenovo V9000, Honor MagicBook and HP EliteDesk equipped with ARM Cortex-A78/A77/A76/A73 CPU, AMD Ryzen 5800H CPU, Intel I9-12900 CPU, and NVIDIA RTX 3000/3070 Ti GPU, AMD Radeon GPU, Intel HD 630 GPU. We fix the maximum frequency on them to ensure consistent performance measurements. We evaluate \sysname on 4 popular browsers: Chrome, Microsoft Edge, Firefox and Opera. All of them support Wasm, but only Chrome has solid WebGPU support.  Without any specific indication, Chrome is the default browser used in the evaluation.  

\begin{figure}
    \centering
    \begin{subfigure}[t]{0.95\linewidth}
        \centering
        \includegraphics[width=\textwidth]{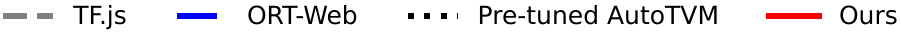}
    \end{subfigure}
    \vfill
    \begin{subfigure}[t]{0.45\linewidth}
        \centering
        \includegraphics[width=\textwidth]{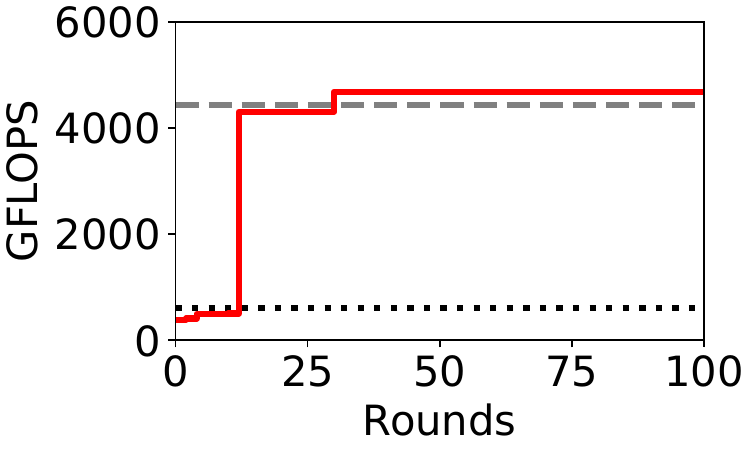}
        \caption{NVIDIA 3070Ti}
    \end{subfigure}
    \begin{subfigure}[t]{0.45\linewidth}  
        \centering 
        \includegraphics[width=\textwidth]{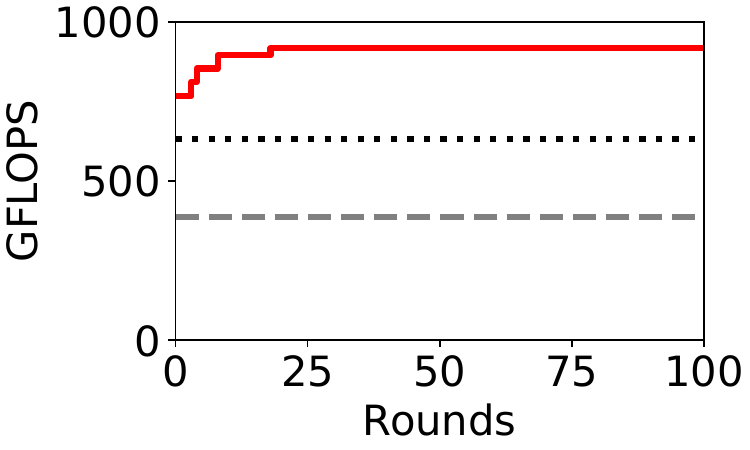}
        \caption{AMD Radeon}
    \end{subfigure}
    \begin{subfigure}[t]{0.45\linewidth}   
        \centering 
        \includegraphics[width=\textwidth]{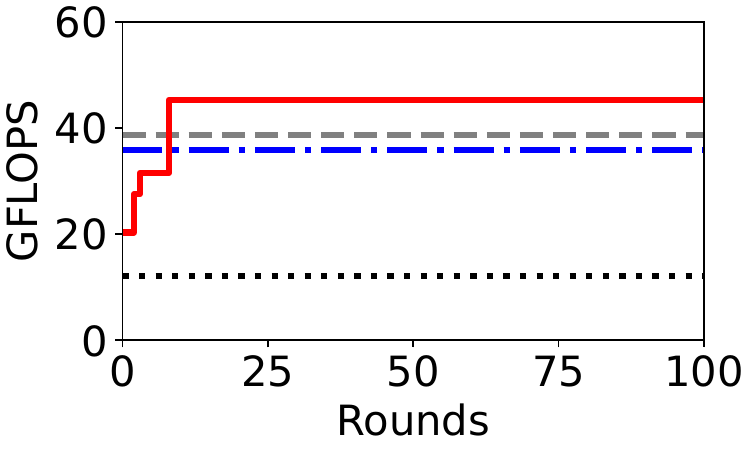}
        \caption{Intel I9 12900H}
    \end{subfigure}
    \begin{subfigure}[t]{0.45\linewidth}   
        \centering 
        \includegraphics[width=\textwidth]{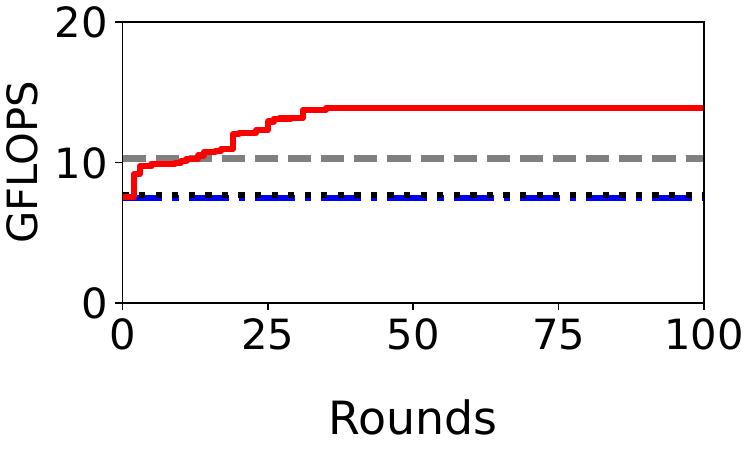}
        \caption{ARM Cortex-A76}
    \end{subfigure}
    \caption{Kernel performance improvements with the JIT kernel optimization rounds on different devices.}
    \label{F.kernel_performance_and_rounds}
\end{figure}

\begin{figure}
    \centering
    \begin{subfigure}[t]{0.45\linewidth}
        \centering
        \includegraphics[width=\linewidth]{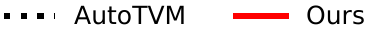}
    \end{subfigure}
    \vfill
    \begin{subfigure}[t]{0.49\linewidth}
        \centering
        \includegraphics[width=\linewidth]{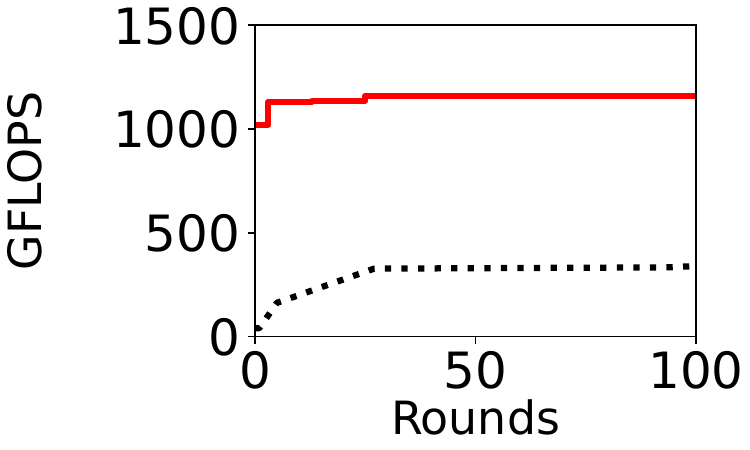}
        \caption{NVIDIA 3050}
    \end{subfigure}
    \begin{subfigure}[t]{0.45\linewidth}   
        \centering 
        \includegraphics[width=\linewidth]{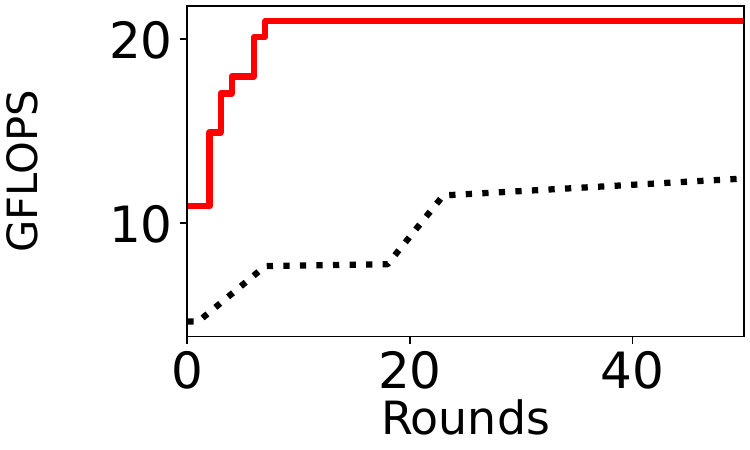}
        \caption{Intel I7 10700}
    \end{subfigure}
    \caption{Kernel performance improvements along with the JIT tuning rounds with \sysname and AutoTVM.}
    \vspace{-0.5em}
    \label{F.customized_kernels}
\end{figure}

\begin{table}[t]
\footnotesize
\caption{Evaluated kernels type and shape.}
\vspace{-1em}
\begin{tabular}{|l|l|l|l|}
\hline
\textbf{ID} & \textbf{Kernel Type} & \textbf{Kernel Size} & \textbf{Model} \\ \hline
K0 & MatMul & M=384,K=768,N=768 & RoBERT \\ \hline
K1 & MatMul & M=640,K=768,N=3072 & GPT-2 \\ \hline
K2 & BatchMatMul & B=12,M=384,K=384,N=64 & BART \\ \hline
K3 & BatchMatMul & B=120,M=64,K=64,N=64 & GPT-2 \\ \hline
\end{tabular}
\label{T.evaluated_kernels}
\end{table}

\textbf{Kernels and models.} We evaluate \sysname on modern transformer models, including RoBERTa~\cite{roberta}, BART~\cite{bart}, GPT-2~\cite{gpt2}, T5~\cite{t5}, and Llama 2 7B~\cite{touvron2023llama}. For the sequence-to-sequence models, such as GPT-2 and T5, we fix the input length at 384 to obtain the comparable results. We also evaluate the performance on typical kernels, picked from the models above, including MatMul and BatchMatMul with different shapes as listed in Table~\ref{T.evaluated_kernels}. 

\begin{figure*}[!ht]
    \centering
    \begin{subfigure}[t]{0.55\textwidth}  
        \centering 
        \includegraphics[width=\linewidth]{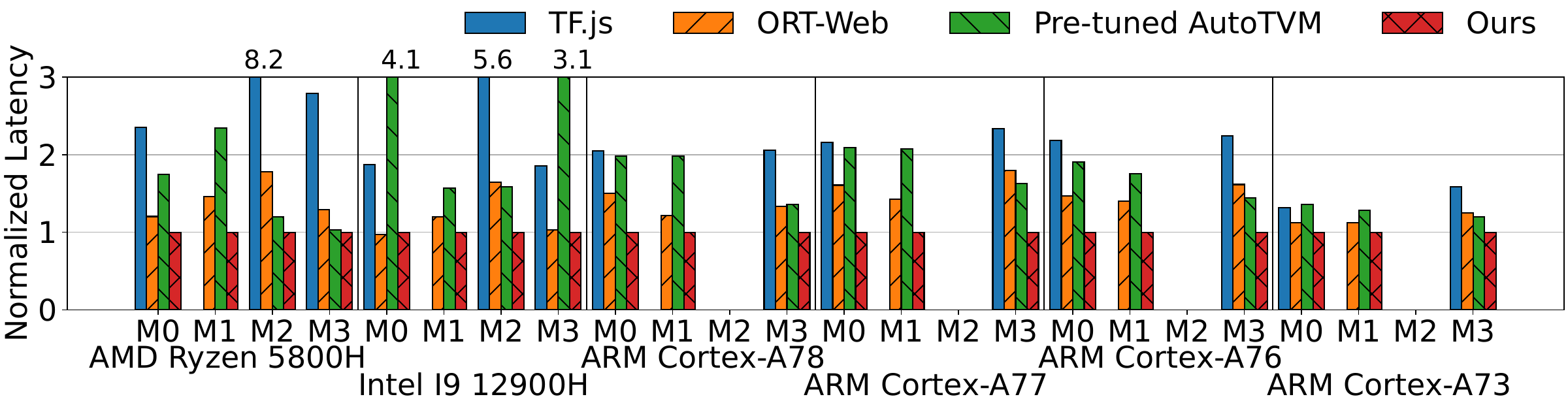}
        \caption{WASM}
    \end{subfigure}
    \begin{subfigure}[t]{0.39\textwidth}
        \centering
        \includegraphics[width=\linewidth]{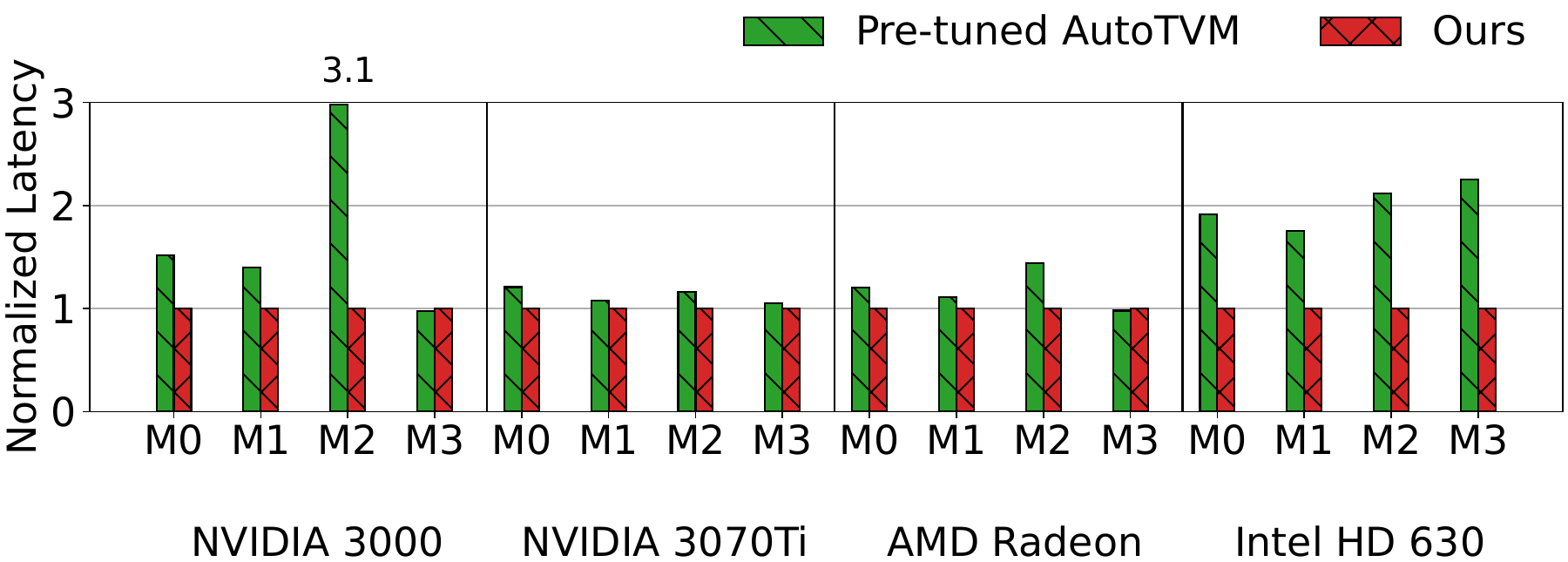}
        \caption{WebGPU}
    \end{subfigure}
    \caption{Model latency executed with TF.js, ORT-web, pre-tuned AutoTVM as well as \sysname.}
    \label{F.model_latency_comparison}
\end{figure*}

\textbf{Baselines}. We compare \sysname with three in-browser DL inference frameworks as baselines, including TF.js (version 3.21.0), ORT-web (version 1.14.0) and pre-tuned AutoTVM~\cite{chen2018tvm}. %
For pre-tuned AutoTVM, we use the default kernel space, search algorithm \ie XGBoost and tuning trails \ie 1000 to generate and tune the kernels ahead-of-time. The pre-tuned target are Intel I7-10700 for Wasm and NVIDIA 3050 for WebGPU. They are excluded in our test devices.

\textbf{Metrics.} We use \code{performance.now()} function, a JaveScript API function to measure the latency of kernels and models running with Wasm, and \code{writeTimestamp} function of WebGPU API to measure the latency on WebGPU. Each kernel and model are evaluated with one warmup and 50 rounds, the averaged latency is reported. A round means one iteration of kernel generation (Fig. \ref{F.tvm_compiling}), starting with a selected candidate configuration and ending with the kernel generation and evaluation. 

To measure memory consumption, we use \code{performance.memory.\\usedJSHeapSize} API function to catch the peak memory usage.

\vspace{-1em}
\subsection{Overall Performance}

\textbf{Kernel performance cross devices.} Fig.~\ref{F.kernel_latency_comparison} demonstrates the latency of tested kernels on CPUs and GPUs, comparing baselines with \sysname. %
On CPUs with Wasm, \sysname achieves an average speedup of 4.59$\times$. On GPUs with WebGPU, it accelerates kernel executions by an average of 2.77$\times$. Specifically, \sysname outperforms TF.js by 5.78$\times$ on Wasm and 1.68$\times$ on WebGPU. When compared to pre-tuned AutoTVM, the speedup is 6.43$\times$ on CPUs and 3.86$\times$ on GPUs. The inference speedup of \sysname is mainly due to efficient kernel tuning for specific hardware, whereas \emph{one-for-all} kernel approaches including TF.js, ORT-Web as well as pre-tuned AutoTVM, fall short in this regard. 

Taking the WebGPU result as an example, pre-tuned AutoTVM behaves much worse on 3070Ti. This is because the pre-tuned AutoTVM kernels are tuned on NVIDIA 3050 (2560 cores), which obviously mismatches with 3070Ti (6144 cores). Therefore, K0 kernel of pre-tuned AutoTVM only activates 50\% warps and utilize only 20\% of L1 cache on 3070Ti.  

\textbf{Kernel performance cross browsers.}
Fig.~\ref{F.kernel_latency_comparison_on_browsers} demonstrates the kernel speedup on four browsers. The actual speedups do have variance, but overall, the speedup pattern for each kernel is similar among browsers. \sysname achieves an average speedup of 5.82$\times$ compared to baselines.%

\textbf{Kernel performance over tuning rounds.}
Fig. \ref{F.kernel_performance_and_rounds} showcases the kernel performance in GFLOPs over JIT tuning rounds on selected CPUs and GPUs. We use the K1 kernel configuration (Table\ref{T.evaluated_kernels}). \sysname attains optimal performance on CPUs with Wasm after 10$\sim$32 tuning rounds, while 25$\sim$40 rounds are needed on GPUs. This can be attributed to the different sizes of web-specific lite spaces. Moreover, our compilation pipeline ensures that each tuning round takes only about 500ms for Wasm and 100ms for WebGPU.

We also compare \sysname with the SOTA \emph{one-for-each} kernel approach. We use AutoTVM and \sysname to tune a kernel for the same hardware, shown in Fig. \ref{F.customized_kernels}.   
The kernel configuration employed is K0 (Table\ref{T.evaluated_kernels}). %
On the Nvidia 3050, \sysname reaches near-peak performance (1159 GFLOPs) within 25 rounds, while AutoTVM lags behind at 350 GFLOPs. On the Intel I7, \sysname finds the best kernel implementation at the 8th round, demonstrating 2.80$\times$ speedup compared to AutoTVM at the same round. AutoTVM needs 1106 additional tuning rounds to achieve its optimal performance.

\begin{figure}
    \centering
    \begin{subfigure}[t]{0.95\linewidth}
        \centering
        \includegraphics[width=\textwidth]{figures/legend_of_kernels.pdf}
    \end{subfigure}
    \vfill
    \begin{subfigure}[t]{0.45\linewidth}
        \centering
        \includegraphics[width=\textwidth]{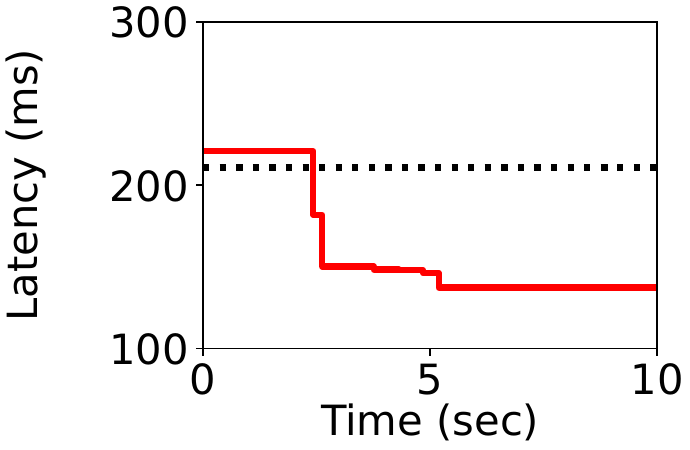}
        \caption{NVIDIA 3000}
    \end{subfigure}
    \begin{subfigure}[t]{0.45\linewidth}  
        \centering 
        \includegraphics[width=\textwidth]{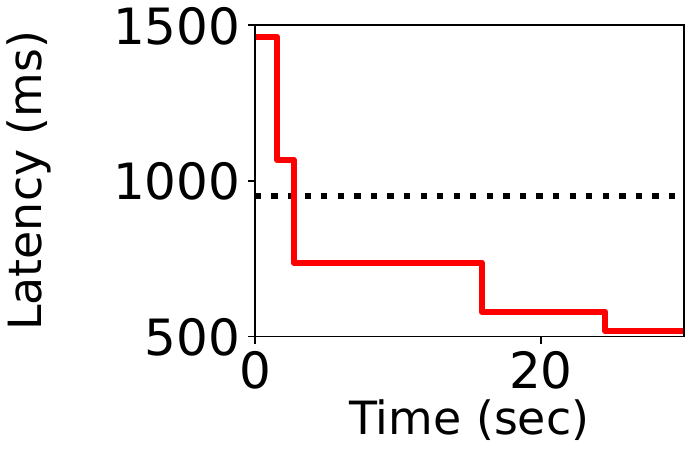}
        \caption{Intel HD 630}
    \end{subfigure}
    \begin{subfigure}[t]{0.45\linewidth}   
        \centering 
        \includegraphics[width=\textwidth]{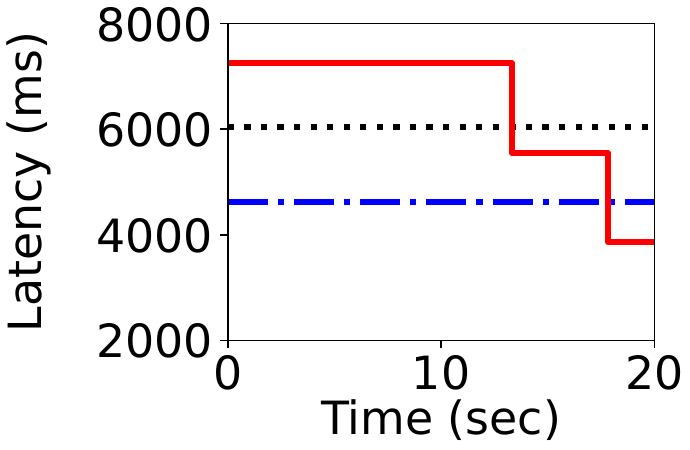}
        \caption{Intel I9 12900H}
    \end{subfigure}
    \begin{subfigure}[t]{0.45\linewidth}   
        \centering 
        \includegraphics[width=\textwidth]{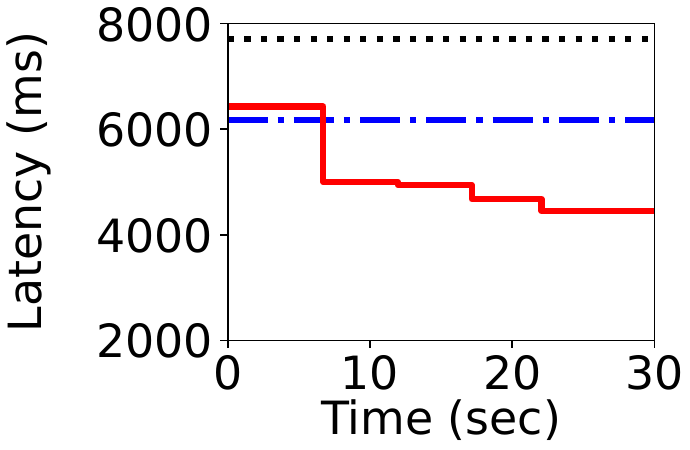}
        \caption{ARM Cortex-A76}
    \end{subfigure}
    \caption{Model performance improvements with the JIT kernel optimization time on different devices.}
    \label{F.model_performance_and_time}
\end{figure}

\textbf{Model performance.} We evaluate the end-to-end model performance achieved by \sysname and other baselines. In Fig.~\ref{F.model_latency_comparison}, we denote RoBERTa as M0, BART as M1, GPT-2 as M2, and T5 as M3. As illustrated, \sysname attains 2.76$\times$ and 1.37$\times$ speedup on average across the tested models compared to TF.js and ORT-Web on CPU with Wasm, respectively. Notably, on the AMD 5800H CPU, \sysname improves by 8.27$\times$ on M3 compared to TF.js, while 5.64$\times$ on Intel I9. M2 (GPT-2) results on two mobile phones are missing because the model size is over the browser memory limitation of Android.   %
As TF.js and ORT-Web cannot support all kernels in the tested models with WebGPU, we do not report their model latency.

Recently, Large Language Models (LLMs) have been gaining prominence. Thus, we evaluate Llama 2 7B model with \sysname under 4-bit quantization. According to our evaluation, on Nvidia RTX 3000 using WebGPU, we achieve the processing speed of 12.16 tokens/s. Compared to the SOTA in-browser LLM, WebLLM~\cite{webllm}, which operates at 8.72 tokens/s, \sysname accelerates LLM in-browser inference by 39.4\%.

We also show the model latency reduction over time, due to the JIT kernel optimization. As shown  in Fig.~\ref{F.model_performance_and_time}, for BART, \sysname takes 5.5 seconds to achieve the optimized latency %
for Nvidia 3000. For CPUs, peak performance is achieved after 17.8s and 22.1s for Intel I9 and ARM A76, respectively.

\begin{table}[t]
\caption{Peak memory consumption (MB) executed with TF.js, ORT-web, pre-tuned AutoTVM as well as \sysname on Chrome.}
\resizebox{\linewidth}{!}
{
\begin{tabular}{|c|c|c|c|c|}
\hline
\textbf{Models} & \textbf{TF.js} & \textbf{ORT-Web} & \textbf{Pre-tuned AutoTVM} & \textbf{Ours}\\
\hline
M0 & 1484 & 1196 & 536 & 550  \\
\hline
M1 & N/A & 1354 & 642 & 656 \\
\hline
M2 & 2293 & 1842 & 636 & 652 \\
\hline
M3 & 398 & 368 & 263 & 268 \\
\hline
\end{tabular}
}
\label{T.memory_comparison}
\end{table}

\textbf{Memory.} We evaluate the peak memory consumption in Table \ref{T.memory_comparison}. The peak memory of \sysname is, on average, 55.7\% and 49.6\% less compared to TF.js and ORT-Web, respectively. Compared with pretuned AutoTVM, \sysname only increases 2.2\% peak memory for JIT kernel optimization. 

\begin{table}[t]
\vspace{-0.5em}
\caption{The compiling latency and achieved kernel latency of \sysname with different optimization passes.}
\resizebox{\linewidth}{!}
{
\begin{tabular}{|c|c|c|}
\hline
& \textbf{Compilation Latency (sec)} & \textbf{Kernel Latency (ms)}\\
\hline
Conventional Compilation & 5.8$\sim$62.9 & 76  \\
\hline
Ours w/o opt. passes & 0.4$\sim$0.5 & 234 \\
\hline
Ours w/ offset load/store & 0.5$\sim$0.6 & 88 \\
\hline

\makecell{Ours w/ offset load/store \\ \& combined instruction} & 0.5$\sim$0.6 & 82 \\
\hline
\makecell{Ours w/ offset load/store \\ \& combined instruction \\ \& load/store to variable} & 0.5$\sim$0.6 & 74 \\
\hline
\end{tabular}
}
\label{T.compiling_latency_comparison}
\end{table}

\begin{table}[t]
\vspace{-0.5em}
\caption{Kernel space of AutoTVM and \sysname.}
\vspace{-1em}
\resizebox{\linewidth}{!}
{
\begin{tabular}{|c|c|c|c|c|}
\hline
\multirow{2}{*}{\textbf{Kernel Type (Size)}} & \multicolumn{2}{c}{\textbf{AutoTVM}} \vline & \multicolumn{2}{c}{\textbf{\sysname}} \vline \\ \cline{2-5}
 & WASM & WebGPU  & WASM & WebGPU \\ \hline
\makecell{MatMul \\ (M=384,K=768,N=768)} & 2,099,520 & 42,768,000  & 10$\sim$32 & 41 \\ \hline
\makecell{BatchMatMul \\ (B=12,M=384,K=384,N=64)} & 2,694,384 & 74,131,200 & 10$\sim$32 & 30 \\ \hline
\end{tabular}
}
\label{T.space_comparison}
\end{table}

\vspace{-0.7em}
\subsection{Ablation Study}
\textbf{Tensor-Web co-designed compilation. }
Table~\ref{T.compiling_latency_comparison} presents the compiling latency and achieved kernel latency for both the baseline and \sysname on AMD 5800H CPU. We use AutoTVM's conventional compilation pipeline as the baseline. For our optimized pipeline, we examine three optimization passes, namely offset load/store, combined instruction, and load/store to variable pass, to assess their individual contributions to the kernel latency reduction. The same kernel implementation is used for all cases.

As demonstrated, our compilation pipeline with all optimizations is over up to 125.8$\times$ faster than the baseline, while maintaining a similar kernel inference latency (76ms and 74ms). Furthermore, our pipeline with three optimization passes results in a significant performance improvement of 166\%, 185\%, and 216\%, respectively, with the compiling overhead increasing by 25\%.

\textbf{Web-Specific lite space. }
Table~\ref{T.space_comparison} compares the size of our lite kernel optimization space and AutoTVM for two typical kernels, MatMul and BatchMatMul.%
Notably, the web-specific lite kernel space size is, on average, around 0.0013\% and 0.000068\% of the AutoTVM space on Wasm and WebGPU, respectively, decreasing search candidates from millions to dozens. In combination with our optimized compilation pipeline, \sysname reduces the overall kernel generation cost from hours to seconds, enabling JIT-powered inference in web browsers.

\textbf{Overhead. }
\sysname enables JIT kernel optimization with minimal overhead. For example, the microbenchmark is a one-time effort executed in the offline stage, taking less than 1 second on a AMD Ryzen 5800H CPU according to our measurements. During JIT inference, kernels are sequentially pushed from the server to devices. The compiled kernel sizes range between 5$\sim$30KB, which does not add a significant burden to the network load. To evaluate the newly arriving kernels, a device typically takes 69$\sim$728ms for most kernels based on our assessment, which is nearly imperceptible.

\section{Related Works}

\textbf{DL kernel generation.} Many works~\cite{chen2018tvm, liang2022romou, zheng2020ansor, antares, tillet2019triton, openxla, iree} focus on automatically searching and generating optimal kernel implementations from a vast space. TVM \cite{chen2018tvm} generate DL kernels based on the space of manual schedule templates and a learned cost model to search for the best kernel implementation. Ansor \cite{zheng2020ansor} and Antares \cite{antares} generates higher-performance DL kernels than TVM without manual schedule templates and reduces the average search time. Romou \cite{liang2022romou} supports new primitives to generate mobile-GPU-friendly DL kernels and accelerates kernel generation through hardware-aware search space pruning.  Although this approach can reduce the space by 99\%, the number of remaining candidates is still on the order of 10K. 
Triton \cite{tillet2019triton} is a DL kernel generator that extends LLVM IR and adds an additional tile-level optimization pass, achieving high DL kernel performance. OpenXLA \cite{openxla} is a ML compiler ecosystem that aims to simplify and accelerate ML development by addressing fragmentation between different ML frameworks and hardware. IREE \cite{iree} is an end-to-end compiler and runtime toolkit specifically for machine learning (ML) models. Compared to \sysname, these works are not for in-browser inference. Besides, the last three works do not really support auto kernel optimization. For example, Triton JIT compiler can only tune CUDA kernels within a space that is designed by experts.

\textbf{In-Browser DL inference.} The emergence of DL frameworks, such as TensorFlow.js \cite{google2023tfjs} and ONNX Runtime Web \cite{microsoft2023ort}, has significantly contributed to making in-browser DL inference a reality. TensorFlow.js from Google supports JavaScript, Wasm, WebGL, and WebGPU. ONNX Runtime Web from Microsoft, facilitates in-browser DL inference by processing models in ONNX format. However, it only supports Wasm and WebGL backends. Transformer.js~\cite{tranformersjs} is recently released to support the transformer model inference in browsers. These frameworks all uses the pre-defined kernels, leading to suboptimal performance across edge devices as discussed in Sec.~\ref{subsec:performance_issues}. \sysname stands out as the first in-browser inference framework that enable JIT compilation, thereby ensuring peak performance across devices. Furthermore, \sysname automatically generates kernels, which allows for the support of the new kernels and models with minimal manual effort.

\section{Conclusion}
In this paper, we present \sysname, the first in-browser inference system that enables JIT optimized kernel generation, supporting Wasm and WebGPU. Our evaluation shows that \sysname accelerates inference by an average of 4.44$\times$ compared to TF.js, ORT-Web, and AutoTVM, while maintaining minimal compilation overhead.

\clearpage
\balance
\bibliographystyle{ACM-Reference-Format}
\bibliography{references}

\end{document}